%% file: ms.tex
\documentclass[letterpaper, 10 pt, journal, twoside]{ieeetran}

\include{preamble}

\title{Model-Based Meta-Reinforcement Learning for Flight with Suspended Payloads}

\author{Suneel Belkhale$^{1}$, Rachel Li$^{1}$, Gregory Kahn$^{1}$, Rowan McAllister$^{1}$, Roberto Calandra$^{2}$, Sergey Levine$^{1}$%
\thanks{Manuscript received: October, 15, 2020; Revised December, 22, 2020;
Accepted January, 14, 2021.}
\thanks{This paper was recommended for publication by Editor Dana Kulic upon evaluation of the Associate Editor and Reviewers' comments.
This research was supported by the National Science Foundation under IIS-1700697 and IIS-1651843, ARL DCIST CRA W911NF-17-2-0181, NASA ESI, the DARPA Assured Autonomy Program, and the Office of Naval Research, as well as support from Google, NVIDIA, and Amazon.
}
\thanks{$^{1}$University of California, Berkeley
{\tt\small \{skbelkhale, rachel\_li, gkahn, rmcallister, svlevine\}@berkeley.edu}}%
\thanks{$^{2} $Facebook AI Research.
{\tt\small rcalandra@fb.com}}%
\thanks{Digital Object Identifier (DOI): 10.1109/LRA.2021.3057046.}
}

\begin{document}

\makeatletter
\let\@oldmaketitle\@maketitle
\renewcommand{\@maketitle}{\@oldmaketitle
    \vspace*{20pt}
    \centering
	\includegraphics[width=0.19\textwidth,trim={1cm 25cm 0 10cm},clip]{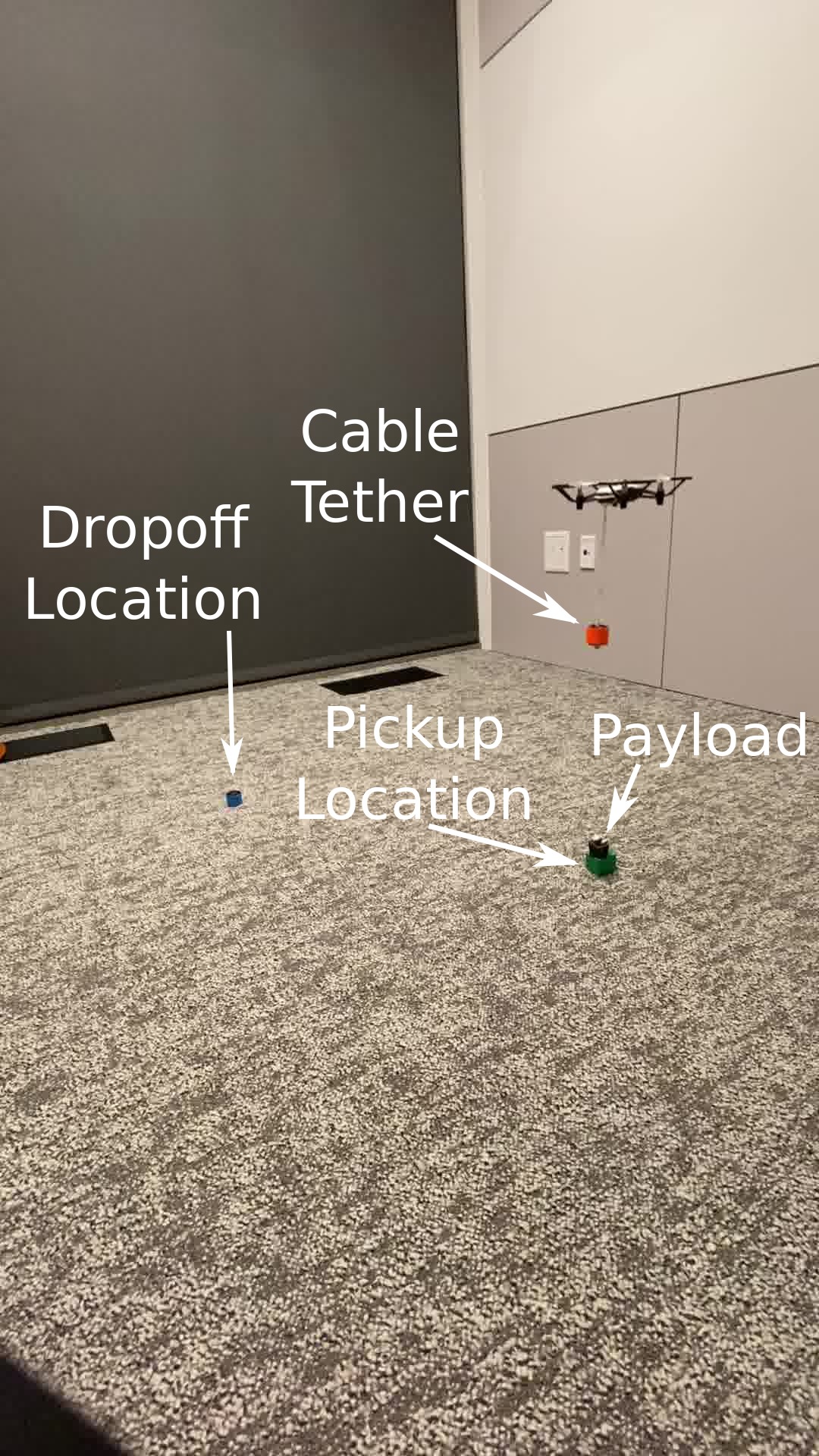}
	\hfill
	\includegraphics[width=0.19\textwidth,trim={1cm 25cm 0 10cm},clip]{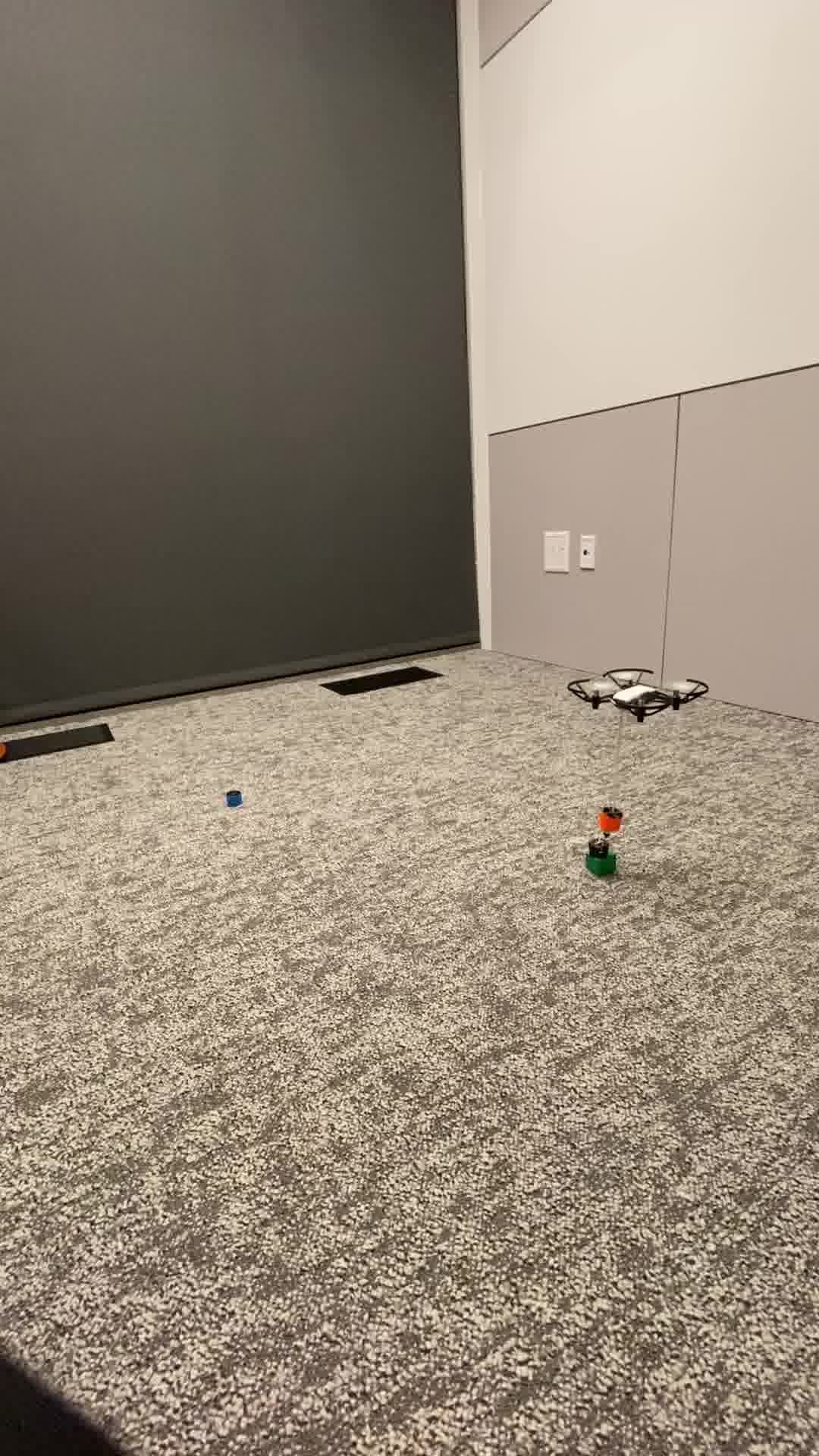}
	\hfill
	\includegraphics[width=0.19\textwidth,trim={1cm 25cm 0 10cm},clip]{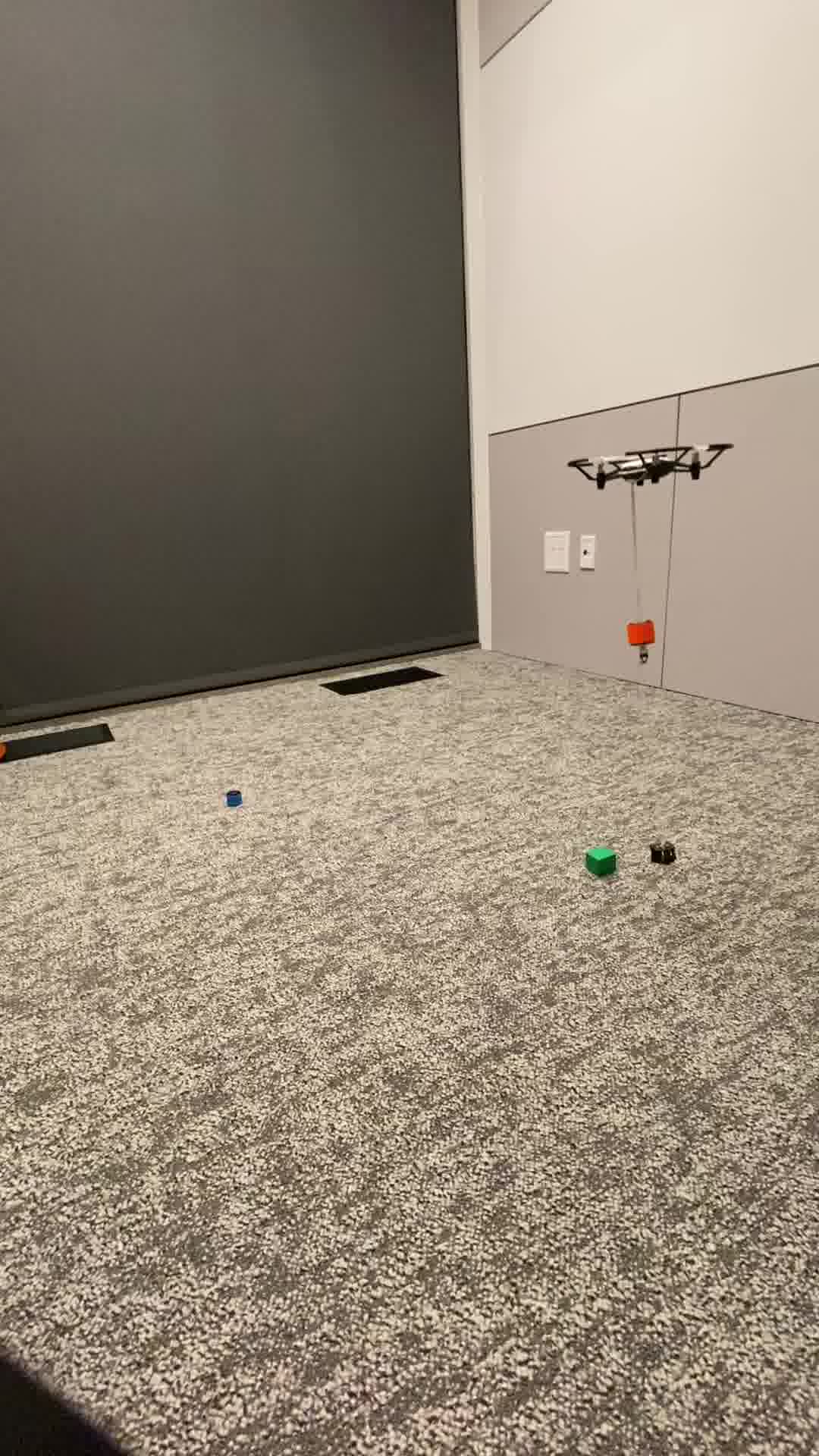}
	\hfill
	\includegraphics[width=0.19\textwidth,trim={1cm 25cm 0 10cm},clip]{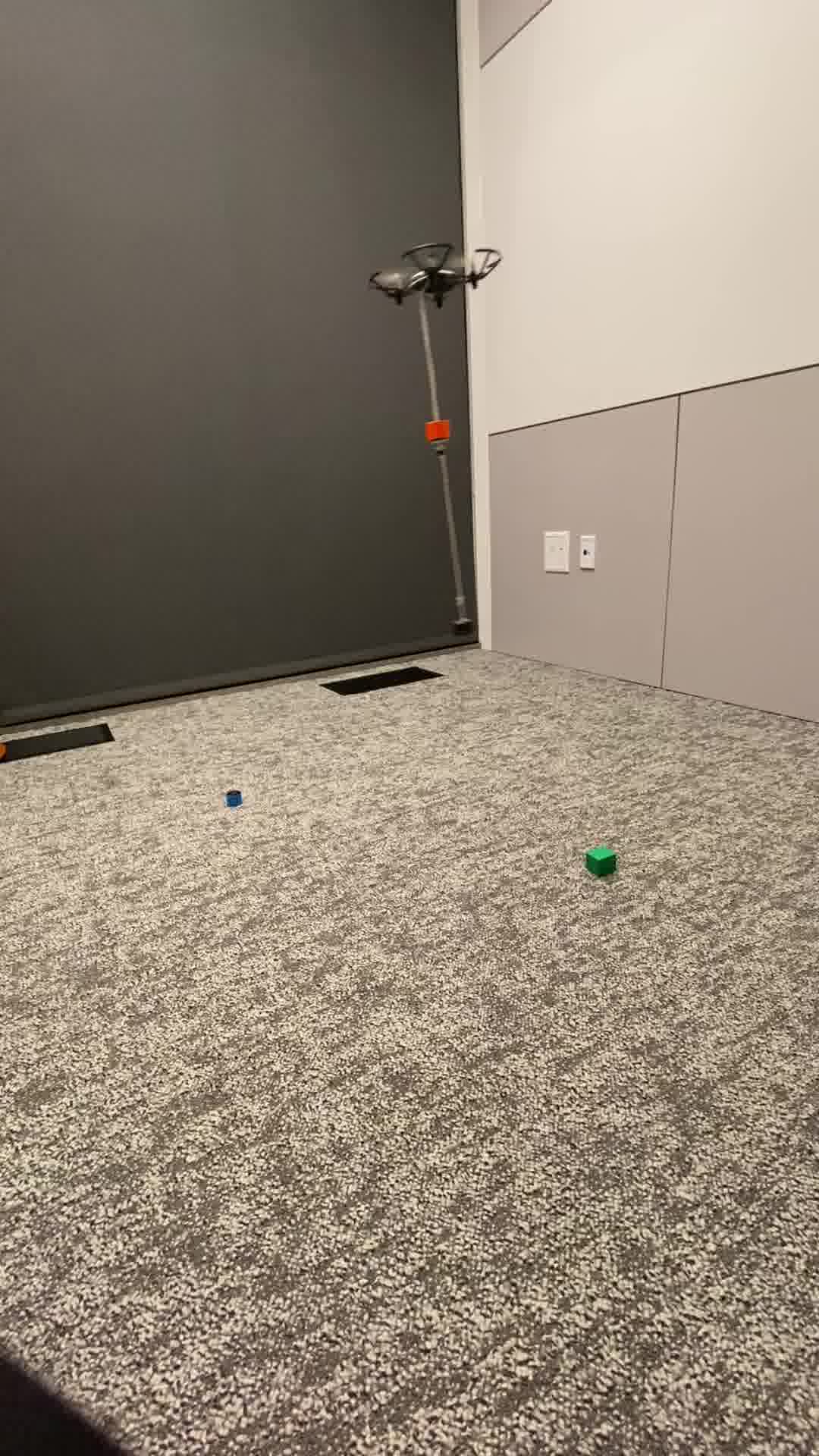}
	\hfill
	\includegraphics[width=0.19\textwidth,trim={1cm 25cm 0 10cm},clip]{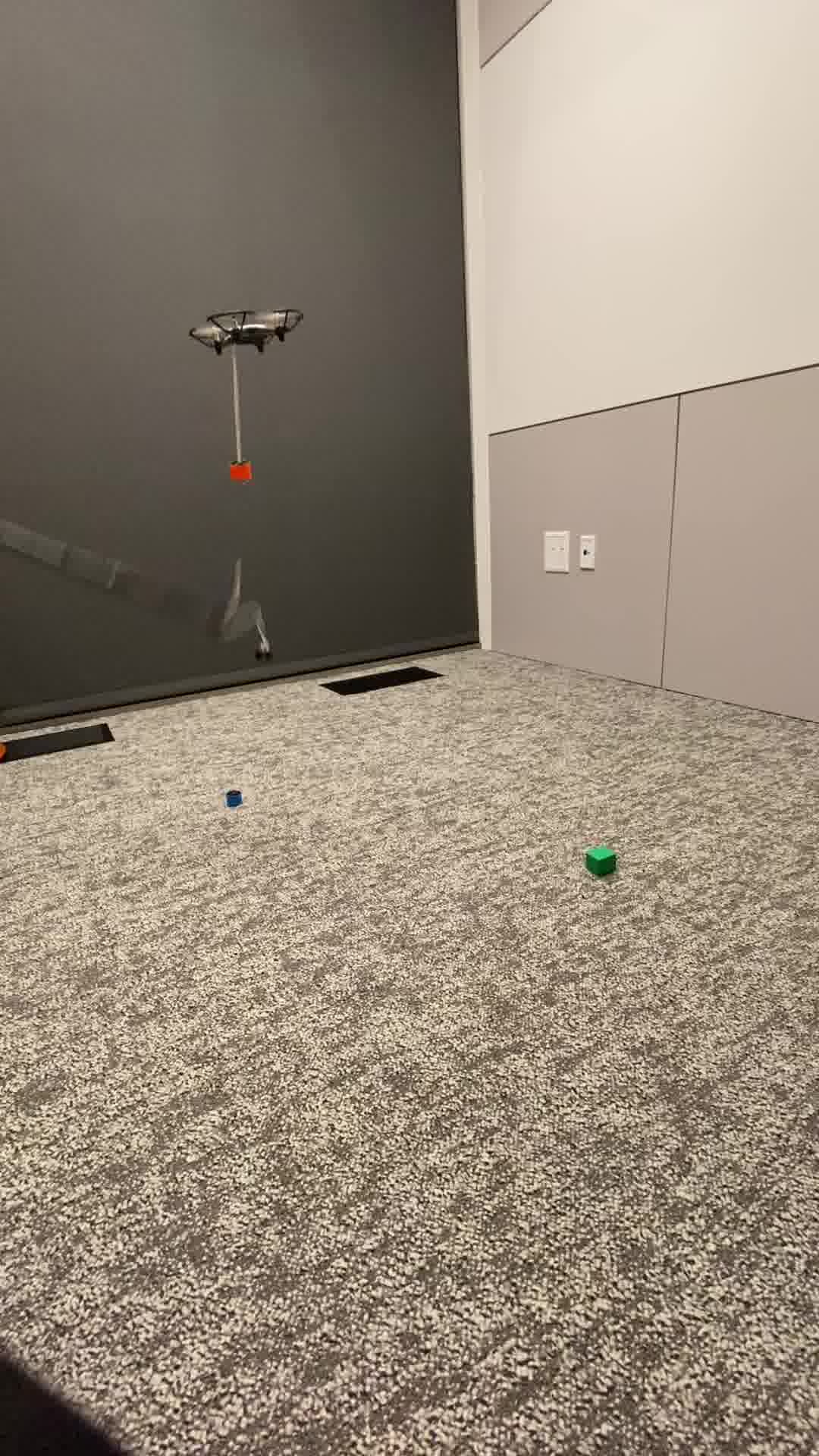}
	{(a)\hspace{93pt}(b)\hspace{93pt}(c)\hspace{93pt}(d)\hspace{93pt}(e)}
	\captionof{figure}{Our meta-reinforcement learning method controlling a quadcopter transporting a suspended payload. This task is challenging since each payload induces different system dynamics, which requires the quadcopter controller to adapt online. The controller learned via our meta-learning approach is able to (a) fly towards the payload, (b) attach the cable tether to the payload using a magnet, (c) take off, (d) fly towards the goal location while adapting to the newly attached payload, and (e) deposit the payload using an external detaching mechanism.}
    \label{fig:results-magnetic}
    \vspace*{-20pt}}
\makeatother
\maketitle



\begin{abstract}
Transporting suspended payloads is challenging for autonomous aerial vehicles because the payload can cause significant and unpredictable changes to the robot's dynamics. These changes can lead to suboptimal flight performance or even catastrophic failure. Although adaptive control and learning-based methods can in principle adapt to changes in these hybrid robot-payload systems, rapid mid-flight adaptation to payloads that have a priori unknown physical properties remains an open problem. We propose a meta-learning approach that ``learns how to learn'' models of altered dynamics within seconds of post-connection flight data. Our experiments demonstrate that our online adaptation approach outperforms non-adaptive methods on a series of challenging suspended payload transportation tasks. Videos and other supplemental material are available on our website: \url{https://sites.google.com/view/meta-rl-for-flight}
\end{abstract}

\begin{IEEEkeywords}
Machine Learning for Robot Control,
Reinforcement Learning,
Probabilistic Inference.
\end{IEEEkeywords}

\IEEEpeerreviewmaketitle


\section{Introduction}
\label{sect:intro}

\IEEEPARstart{C}{onsider}
the task illustrated in \fig{results-magnetic}: the quadcopter must maneuver such that the magnet (red) at the end of the tether picks up the payload (green), then lift the payload, and fly such that this payload follows a desired trajectory. While the dynamics of the quadcopter may be well-characterized, and system identification methods could accurately identify its parameters, the complex interaction between the magnetic gripper and the payload are unlikely to be represented accurately by hand-designed models. Even more unpredictable is the effect of the payload on the dynamics of the quadcopter when the payload is lifted off the ground. For example, a payload attached with a shorter cable will oscillate faster compared to one attached with a longer cable. Because the robot will be picking up and dropping off various \textit{a priori} unknown payloads, the robot must rapidly adapt to the new dynamics to remain in flight. Learning can offer us a powerful tool for handling complex interactions, such as those between the magnetic gripper and the payload. Conventional learning-based methods, however, typically require a large amount of data to learn accurate models, and therefore may be slow to adapt. The payload adaptation task illustrates the need for fast adaptation: the robot must very quickly determine the payload parameters, and then adjust its motor commands accordingly. To address the challenge of rapid \textit{online} adaptation, we propose an approach based on meta-learning. In our meta-learning formulation, we explicitly train the model for fast adaptation to scenarios with new dynamics, such as the task in \fig{results-magnetic}.

Our algorithm can be viewed as a model-based meta-reinforcement learning method: we learn a predictive dynamics model, represented by a deep neural network, which is augmented with stochastic latent variables that represent the unknown factors of variation in the environment and task. The model is trained with different payload masses and tether lengths, and uses variational inference to estimate the corresponding posterior distribution over these latent variables. This training procedure enables the model to adapt to new payloads at test-time by inferring the posterior distribution over the latent variables. 
Our novel contribution is in leveraging neural network dynamics models in conjunction with meta-learning for the task of controlling an aerial robot with a suspended payload.

In the experiments, we demonstrate that our method enables a quadcopter to plan and execute trajectories that follow desired payload trajectories, drop off these payloads at designated locations, and even pick up new payloads with a magnetic gripper. To our knowledge, this is the first meta-learning approach demonstrated on a real-world quadcopter using only real-world training data that successfully shows improvement in closed-loop performance compared to non-adaptive methods for suspended payload transportation.


\section{Related Work}
\label{sect:related}

Prior work on control for aerial vehicles has demonstrated impressive performance and agility, such as enabling aerial vehicles to navigate between small openings~\cite{mellinger2012trajectory}, perform aerobatics~\cite{lupashin2010simple}, and avoid obstacles~\cite{richter2016polynomial}. These approaches have also enabled aerial vehicles to aggressively control suspended payloads~\cite{tang2015mixed,tang2018aggressive}. These methods typically rely on manual system identification, in which the equations of motion are derived and the physical parameters are estimated for both the aerial vehicle~\cite{mahony2012multirotor,zhang2014survey} and the suspended payload~\cite{tang2015mixed,tang2018aggressive}. Although these approaches have successfully enabled controlled flight of the hybrid system, they require \textit{a priori} knowledge of the system, such as the payload mass and tether length~\citep{Faust2017}. When such parameters cannot be identified in advance, alternative techniques are required.

Many approaches overcome the limitations of manual system identification by performing automated system identification, in which certain parameters are automatically adapted online according to a specified error metric~\cite{slotine1987adaptive,ioannou2006adaptive,kupcsik2013data}. However, the principal drawback of manual system identification---the reliance on domain knowledge for the equations of motion---still remains. 
While certain rigid-body robotic systems are easily identified, more complex phenomena, such as friction, contacts, deformations, and turbulence, may have no known analytic equations (or known solutions). In suspended payload control, adaptive model-based controllers could require \textit{a priori} knowledge of the quadrotor dynamics and suspended payload equations, 3d state estimation of both the quadrotor and suspended payload, and camera parameters in order to perform state estimation. In contrast, our data-driven method only requires the pixel location of the payload and the quadrotor’s commanded actions to control and adapt to the suspended payload's dynamics.

Prior work has also proposed end-to-end learning-based approaches that learn from raw data, such as value-based methods which estimate cumulative rewards~\cite{watkins1992q} or policy gradient methods that directly learn a control policy~\cite{williams1992simple}. Although these model-free approaches have been applied to various tasks~\cite{mnih2013playing,schulman2015trust}, including for robots~\cite{kalashnikov2018qt}, the learning process generally takes hours or even days, making it poorly suited for safety-critical and resource-constrained quadcopters.

Model-based reinforcement learning (MBRL) can provide better sample efficiency, while retaining the benefits of end-to-end learning~\cite{deisenroth2011pilco,gal2016improving,nagabandi2018neural,chua2018deep}. With these methods, a dynamics model is learned from data and then either used by a model-based controller or to train a control policy. Although MBRL has successfully learned to control complex systems such as quadcopters~\cite{Bansal2016,Lambert2019Low}, most MBRL methods are designed to model a single task with unchanging dynamics, and therefore do not adapt to rapid online changes in the system dynamics.

One approach to enable rapid adaptation to time-varying dynamical systems is \textit{meta-learning}, which is a framework for \textit{learning how to learn} that typically involves fine-tuning of a model's parameters \cite{finn2017model,Harrison2019Control,nagabandi2019learning} or input variables \cite{perez2018efficient,saemundsson2018meta}. There has been prior work on model-based meta-learning for quadcopters. \citet{o2019meta} used the MAML~\citep{finn2017model} algorithm for adapting a drone's internal dynamics model in the presence of wind. Although they demonstrated the meta-learning algorithm improved the model's accuracy, the resulting adapted model did not improve the performance of the closed-loop controller. In contrast, we demonstrate that our meta-learning approach does improve performance of the model-based controller. 
\citet{nagabandi2019learning} and \citet{kaushik2020fast} also explored meta-learning for online adaptation in MBRL for a legged robot, demonstrating improved closed-loop controller performance with the adapted model. Our work focuses on suspended payload manipulation with quadcopters, which presents an especially prominent challenge due to the need for rapid adaptation in order to cope with sudden dynamics changes when picking up payloads.


\input{tikz/flow.tex}

\section{Preliminaries}
\label{sect:preliminaries}

We first introduce our notation, problem formulation, and preliminaries on model-based reinforcement learning (MBRL) that our meta-learning algorithm builds upon. We represent the hybrid robot-environment system as a Markov decision process, with continuous robot-environment state $\bs\in\mathbb{R}^{d_\bs}$, continuous robot action $\ba\in\mathbb{R}^{d_\ba}$, and discrete time steps $t$. The state evolves each time step according to an unknown stochastic function $\bs_{t+1} \sim p(\bs_{t+1} | \bs_{t}, \ba_{t})$. We consider $K$ tasks $\{\task_1.,...,\task_K\}$. In each task, the robot's objective is to execute actions that maximize the expected sum of future rewards $r(\bs_t, \ba_t)\in\mathbb{R}$ over the task's finite time horizon $T$.

We approach this problem using the framework of model-based reinforcement learning, which estimates the underlying dynamics from data, with minimal prior knowledge of the dynamics of the system. We can train a dynamics model $p_\theta(\bs_{t+1} | \bs_{t}, \ba_{t})$ with parameters $\theta$ by collecting data in the real world, which we can view as sampling ``ground truth'' tuples $(\bs_t, \ba_t, \bs_{t+1})$. By collecting a sufficient amount of empirical data $\datatrain = \{(\bs_0, \ba_0, \bs_1), (\bs_1, \ba_1, \bs_2), ...\}$, we can train the parameters $\theta$ of the dynamics model via maximum likelihood
\begin{align}
\theta^* \;&=\; \argmax_{\theta} \; p(\datatrain | \theta) \nonumber \\
\;&=\; \argmax_{\theta} \sum_{(\bs_t,\ba_t,\bs_{t+1}) \in \datatrain} \!\!\!\! \log p_\theta(\bs_{t+1}|\bs_{t},\ba_{t})\,. \label{eq:train}
\end{align}
To instantiate this method, we extend the PETS algorithm~\cite{chua2018deep}, which has previously been shown to handle expressive neural network dynamics models and attain good sample efficiency and final performance. PETS uses an ensemble of neural network models, each parameterizing a Gaussian distribution on $\bs_{t+1}$ conditioned on both $\bs_t$ and $\ba_t$. The learned dynamics model is used to plan and execute actions via model predictive control (MPC)~\cite{mpc,kamthe2017data,nagabandi2018neural}. MPC uses the dynamics model to predict into the future, and selects the action sequence with the highest predicted reward
\begin{equation}
\ba_{t}^* \;=\; \argmax_{\ba_{t}} \left[ \max_{\ba_{t+1:t+H}} \sum_{\tau=t}^{t+H} \mathbb{E}_{\bs_\tau\sim p_\theta}\left[r(\bs_\tau, \ba_\tau)\right]\right]\,, \label{eq:mpc}
\end{equation}
in which $\bs_\tau$ is recursively sampled from the learned dynamics model: $\bs_{\tau+1}\sim p_\theta(\bs_{\tau+1} | \bs_{\tau}, \ba_{\tau})$, initialized at $\bs_{\tau}\!\leftarrow\!\bs_t$. Once this optimization is solved, only the first action $\ba_t^*$ is executed. 
A summary of this MBRL framework is provided in \algo{brief}, and we refer the reader to \citet{chua2018deep} for additional details.

\begin{algorithm}[h]
    \small
    \begin{algorithmic}[1]
    \floatname{algorithm}{Procedure}
     \STATE{Initialize dynamics model $p_\theta$ with random parameters $\theta$}
     \WHILE{not done}
        \STATE{Get current state $\bs_t$}
        \STATE{Solve for action $\ba^*_t$ given $p_{\theta^*}$ and $\bs_t$ using MPC \hfill $\triangleright$ see \eqref{eq:mpc}}
        \STATE{Execute action $\ba^*_{t}$}
        \STATE{Record outcome: $\datatrain \leftarrow \datatrain \cup \{\bs_{t}, \ba^*_{t}, \bs_{t+1}\}$}   
        \STATE{Train dynamics model $p_\theta$ using $\datatrain$ \hfill $\triangleright$ see \eqref{eq:train}\;}
     \ENDWHILE
    \end{algorithmic}
    \caption{Model-Based Reinforcement Learning}
    \label{algo:brief}
\end{algorithm}
\vspace{-12pt}


\section{Model-Based Meta-Learning for\\Quadcopter Payload Transport}
\label{sect:method}

Our goal is to enable a quadcopter to precisely control a wide variety of payloads without prior knowledge of the payload's physical properties. The primary challenge is that the interaction between quadcopter actions and the suspended payload state varies based on the type of payload, and these variations in dynamics are difficult to identify and model \textit{a priori}.
Although prior work on MBRL has been able to learn to control complex systems, MBRL is unable to account for factors of variation that are not accounted for in the state $\bs$.
We approach this problem of accounting for \textit{a priori} unspecified factors of variation through the lens of meta-learning, in which we learn a model that is explicitly trained to adapt online.

The quadcopter's objective is to pick up and transport a suspended payload along a specified path to reach a goal location (\fig{results-magnetic}). First, the quadcopter must fly to the location of the payload (\fig{results-magnetic}a), attach itself to the payload using a suspended cable (\fig{results-magnetic}b), and then lift the payload off the ground (\fig{results-magnetic}c). The magnetic gripper is at the end of a tether, so its dynamics are themselves complex and assumed to be unknown before training. As soon as the quadcopter takes off with the payload, the quadcopter's dynamics change drastically, and therefore online adaption is critical. As the quadcopter flies with the payload towards the goal location (\fig{results-magnetic}d), our method continuously adapts to the new payload by updating and improving its dynamics model estimate in real time. The adaptive model improves the performance of the MPC controller, which enables the quadcopter to reach the goal destination and release the payload (\fig{results-magnetic}e). The quadcopter is then able to continue transporting other payloads by adapting online to each new payload it transports. 

\subsection{Data Collection}\label{sect:datacollection}

We first collect data by manually piloting the quadcopter (\fig{flow}, left) along random paths for each of the $K$ suspended payloads, though any off-policy data collection method that visits a diverse number of state and action sequences could also be used.
We save all the data into a single dataset $\datatrain$, consisting of $K$ separate datasets $\datatrain\doteq\setK{\datatrain}\doteq\{\datatrain_1,...,\datatrain_K\}$, one per payload task.

\subsection{Model Training with Known Dynamics Variables}
\label{sect:method-known-vars}

In this section, we consider the case in which we know all the ``factors of variation'' in the dynamics across tasks, represented explicitly as a ``dynamics variable'' $\bz_k\in\mathbb{R}^{d_\bz}$ that is known at training time, but unknown at test-time (deployment). For example, we might know that the only source of variation is the tether length $L$, and therefore we can specify $\bz_k\!\leftarrow\!L_k\,\forall\,k$ at training time. We can then learn a single dynamics model $p_\theta$ across all tasks by using $\bz_k$ as an auxiliary input to PETS~\cite{chua2018deep}:
\begin{equation}
\bs_{t+1} \;\sim\; p_\theta(\bs_{t+1} | \bs_{t}, \ba_{t}, \bz_k).\label{eq:modelsample}
\end{equation}
Having $\bz_k$ as an auxiliary input is necessary for accurate modelling because the factors of variation that affect the payload's dynamics, such as the tether length, are not present in the state $\bs$, which only tracks the position of the tether end point. 
The combination of both $\bs_t$ and $\bz_t$ is more complete representation of the hybrid robot-payload system state, which enables more accurate predictions.

Training is therefore analogous to \eqref{eq:train}, but with an additional conditioning on $\bzall\doteq[\bz_1,...,\bz_K]$:
\begin{align}
    \theta^* 
    &\;\doteq\; \argmax_\theta \, \log p(\datatrain|\bzall, \theta) \nonumber \\[-5pt]
    &\;=\; \argmax_\theta \sum_{k=1}^{\text{K}} \; \sum_{(\bs_t,\ba_t,\bs_{t+1}) \in \datatrain_k} \hspace{-5mm} \log p_\theta(\bs_{t+1}|\bs_{t},\ba_{t}, \bz_k)\,. \label{eq:model}
\end{align}
The variables in this training process can be summarized in the graphical model shown in \fig{graphKnownZ}, in which every variable is observed except for the ``true'' model parameters $\theta$, which we infer approximately as $\theta^*$ using maximum likelihood estimation in \eqref{eq:model}.

\subsection{Meta-Training with Latent Dynamics Variables}
\label{sect:method-latent-vars}

The formulation in \sect{method-known-vars} requires knowing the dynamics variables $\bzall$ at training time. This is a significant assumption because not only does it require domain knowledge to \textit{identify} all possible factors of variation, but also that we can \textit{measure} each factor at training time.

To remove this assumption, we now present a more general training procedure that \textit{infers} the dynamics variables $\bzall$ and the model parameters $\theta$ \textit{jointly}, as shown by \fig{graphUnknownZ}, without needing to know the semantics or values of $\bzall$. We begin by placing a prior over $\bzall\sim p(\bzall)=\mathcal{N}(0,I)$, and then jointly infer the posterior $p(\theta, \bzall|\setK{\datatrain})$. We refer to this as \textit{meta-training}, summarized graphically in \fig{graphUnknownZ} and shown in the broader algorithm flow diagram in \fig{flow} (center).

\input{tikz/graphs.tex}

Unfortunately, inferring $p(\theta, \bzall|\setK{\datatrain})$ exactly is computationally intractable. We therefore approximate this distribution with an approximate---but tractable---variational posterior~\citep{jordan1999introduction}, which we choose to be a Gaussian with diagonal covariance, factored over tasks,
\begin{align}
    q_{\phi_k}(\bz_k) \;=\; \mathcal N({\mu_{k}}, {\Sigma_{k}}) \;\approx\; p(\bz_k|\datatrain)\hspace{5mm}\forall\;k\in[K]\,,
\end{align}
and parameterized by $\phi_k\doteq\{\mu_K,\Sigma_k\}$.
Our meta-learning training objective is to again maximize the likelihood of the full dataset $\datatrain=\setK{\datatrain}$, analogous to Equation~\eqref{eq:model}. The only difference to \sect{method-known-vars} is that we must (approximately) marginalize out $\bzall$ because it is unknown:
{\small
\begin{align}
    &\log p(\datatrain|\theta) \;=\; \log \int_{\bzall} p(\datatrain | \bzall, \theta) p(\bzall) \text{d}\bzall \nonumber \\[-10pt]
    &\quad=\; \sum_{k=1}^{K} \log \mathbb{E}_{\bz_k \sim q_{\phi_k}} p(\datatrain | \bz_k, \theta) \cdot p(\bz_k) / q_{\phi_{k}}(\bz_k) \nonumber \\[-2pt]
    &\quad\geq\; \sum_{k=1} \mathbb{E}_{\bz_k\sim {q_{\phi_k}}} \hspace{-16mm} \sum_{\qquad\qquad\;\;(\bs_t,\ba_t,\bs_{t+1})\in \datatrain_k} \hspace{-15mm} \log p_{\theta}(\bs_{t+1}|\bs_{t},\ba_{t}, \bz_k) \!-\! \text{KL}({q_{\phi_k}}\!(\bz_k) || p(\bz_k)) \nonumber \\ 
    &\quad\doteq\; \text{ELBO}(\datatrain | \theta, \phi_{1:K})\,.
     \label{eq:elbo}
\end{align}
}
The evidence lower bound (ELBO) above is a computationally tractable approximate to $\log p(\datatrain|\theta)$.
For additional details on variational inference, we refer the reader to \citet{bishop2006pattern}.

Our meta-training algorithm then optimizes both $\theta$ and the variational parameters $\phi_{1:K}$ of each task with respect to the evidence lower bound
\begin{align}
    \theta^* &\;\doteq\; \argmax_{\theta} \,\, \max_{\phi_{1:K}} \,\, \text{ELBO}(\datatrain | \theta, \phi_{1:K})\,. \label{eq:thetaphi}
\end{align}
Note that $\theta^*$ will be used at test time, while the learned variational parameters $\phi_{1:K}$ will not be used at test time because the test task can be different from the training tasks. 

\subsection{Test-Time Task Inference}
\label{sect:method-test-time}

At test time, the robot must adapt online to the new task---such as a different type of payload---by inferring the unknown dynamics variables $\bztest$ in order to improve the learned dynamics model $p_{\theta^*}$ and the resulting MPC planner. Inference is performed by accumulating transitions $(\bs_t, \ba_t, \bs_{t+1})$ into $\datatest$, and using this data and the meta-trained model parameters $\theta^*$ to infer the current value of $\bztest$ in real time, as seen in the right side of \fig{flow}. A summary of the variables involved in the inference task is given by \fig{graphTest}.

Similarly to \sect{method-latent-vars}, exact inference is intractable, and we therefore use a variational approximation for $\bztest$:
\begin{align}
    q_{\phitest}(\bztest) \;=\; \mathcal N({\mu^{\text{test}}}, {\Sigma^{\text{test}}}) \;\approx\; p(\bztest|\datatest)\,,
\end{align}
parameterized by $\phitest\doteq\{\mu^{\text{test}},\Sigma^{\text{test}}\}$. 
Regardless of training regime (\sect{method-known-vars} or \sect{method-latent-vars}), inferring $\bztest$ uses the same procedure outline below.

To infer the relevant effects that our test-time payload is having on our system, we again use variational inference to optimize $\phitest$ such that the approximate distribution $q_{\phitest}(\bztest)$ is close to the true distribution $p(\bztest|\datatest)$, measured by the Kullback-Leibler divergence:
{\small
\begin{align}
&\phi^*
\doteq \argmax_{\phi} \; -\text{KL}({q_{\phi}}(\bztest) |\!| p(\bztest|\datatest, \theta^*)) \nonumber \\[-5pt]
&= \argmax_{\phi}  \mathbb{E}_{\bztest \sim q_{\phi}} \log p(\bztest|\datatest, \theta^*) \!-\! \log q_{\phi}(\bztest) \nonumber \\[-5pt]
&= \argmax_{\phi}  \mathbb{E}_{\bztest \sim q_{\phi}} \log p(\datatest|\bztest, \theta^*) \!-\! \log q_{\phi}(\bztest) \nonumber \!+\! \log p(\bztest) \nonumber \\[-5pt]
&= \argmax_{\phi}  \mathbb{E}_{\bztest \sim q_{\phi}} \hspace{-15mm} \sum_{\qquad\qquad\;\;(\bs_t,\ba_t,\bs_{t+1}) \in \datatest} \hspace{-15mm} \log p_{\theta^*}\!(\!\bs_{t+1}|\bs_{t},\!\ba_{t},\!\bztest) \!-\! \text{KL}(\!q_{\phi}(\bztest) |\!| p(\bztest)\!) \nonumber \\
&= \argmax_{\phi} \; \text{ELBO}(\datatest|\theta^*,\phi)\,. \label{eq:latent}
\end{align}
}
Note the objective \eqref{eq:latent} corresponds to the test-time ELBO of $\datatest$, analogous to training-time ELBO of $\datatrain$ \eqref{eq:elbo}. Thus \eqref{eq:latent} scores how well $\phi$ describes the new data $\datatest$, under our variational constraint that $q$ is assumed to be Gaussian. Since $\theta^*$ was already inferred at training time, we treat it as a constant during this test-time optimization. Equation \eqref{eq:latent} is tractable to optimize, and therefore at test time we perform gradient descent online in order to learn ${\phitest}$ and therefore improve the predictions of our learned dynamics model. 

\subsection{Method Summary}\label{sect:method-summary}

A summary of the full training and test-time procedure is provided in both \fig{flow} and Algorithm~\ref{alg:full}.
During the training phase, a human pilot gathers data for $K$ different tasks consisting of suspended payloads with different dynamics. 
During flight, tuples $\{\bs_t, \ba_t, \bs_{t+1}\}$ are recorded into the corresponding task dataset, as well as the dynamics variable $\bz_k$ if it is known (\sect{method-known-vars}). We then train the dynamics model $p_{\theta^*}$ using the dataset $\datatrain$ via \eqref{eq:thetaphi}.

At test time, we initialize $q_{\phitest}(\bztest)$ to be the prior $\mathcal{N}(0,I)$ and the quadcopter begins to transport payloads with \textit{a priori} unknown physical properties $\bztest$. At each time step, we solve for the optimal action $\ba_t^*$ given $p_{\theta^*}$ and the current estimate of $\bztest$ using the MPC planner in \eqref{eq:modelsample}. The quadcopter executes the resulting action and records the observed transition $\{\bs_t, \ba_t^*, \bs_{t+1}\}$ into the test dataset $\datatest$. We then adapt the latent variable online by inferring $q_{\phi^*}(\bztest)$ using $\datatest$. The quadcopter continues to plan, execute, and adapt online until the payload transportation task is complete.

\subsection{Method Implementation}\label{sect:method-implementation}

The quadcopter we use is the DJI Tello (\fig{results-magnetic}), which enables rapid experimentation for suspended payload control due to its small \SI{98 x 93 x 41}{\milli\meter} size, light \SI{80}{\gram} weight, long 13 minute battery life, and powerful motors. In our tasks, 3D printed payloads weighing between 10--15 grams are attached to the Tello via a string. Our experiments vary primarily the string length between 18--30cm long 
since the dynamics are more sensitive to string length than mass. 
We found that this range of string lengths exhibits a larger variation in dynamics while staying in the field of view of our external camera and not interfering with onboard altitude estimation.

During data collection, we record the controls (actions) and the location of the payload, which we track with an externally mounted RGB camera using OpenCV~\citep{bradski2008learning}. The recorded actions are Cartesian velocity commands $\ba \in \mathbb{R}^3$ and the recorded states are the pixel location and size of the payload $\bs \in \mathbb{R}^3$, which are saved every control step into the corresponding dataset in $\datatrain$. 
In our comparative experiments, the final dataset $\datatrain$ consisted of approximately 16,000 data points (1.1 hours of flight) split between 18cm and 30cm, which were then used by our meta-learning for model-based reinforcement learning algorithm.

We instantiate the dynamics model as a neural network consisting of four fully-connected hidden layers of size 200 with swish activations. The model was trained using the Adam optimizer with learning rate 0.001. We used 95\% of the data for training and 5\% as holdout. The model chosen for evaluation was the one which obtained the lowest loss on the holdout data.
MPC is run on an external laptop, with a time horizon of 5 steps, and we used the cross entropy method \citep{botev2013cross} to optimize, with a sample size 50, selecting 10 elite samples, and 3 iterations. This computation takes 50-100ms, so we selected our control frequency to be 4Hz for both training and test time, to allow the remaining 150-200ms for latent variable inference at each step.
We adapted code from a PyTorch implementation of PETS~\cite{chua2018deep} found \url{https://github.com/quanvuong/handful-of-trials-pytorch}.

\renewcommand\algorithmiccomment[1]{\hfill $\triangleright$ #1}
\begin{algorithm}[t]
    \small
    \begin{algorithmic}[1]
    \floatname{algorithm}{Procedure}
     \STATE{\textit{// Training Phase}}
     \FOR {Task $k = 1$ to $K$}
     \FOR {Time $t = 0$ to $T$}
     \STATE{Execute action $\ba_{t}$ from human pilot}
     \IF[case \sect{method-known-vars}]{$\bz_k$ is known}
     \STATE{Record outcome: $\datatrain \leftarrow \datatrain \cup \{\bs_{t}, \ba_{t}, \bs_{t+1}, \bz_k\}$}
     \ELSE[case \sect{method-latent-vars}]
     \STATE{Record outcome: $\datatrain \leftarrow \datatrain \cup \{\bs_{t}, \ba_{t}, \bs_{t+1}\}$}
     \ENDIF
     \ENDFOR
     \ENDFOR
     \STATE{Train dynamics model $p_{\theta^*}$ given $\datatrain$ \hfill $\triangleright$ see \eqref{eq:thetaphi}}
     \vspace{2mm}
     \STATE{\textit{// Test Phase}}
     \STATE{Initialize variational parameters: $\phi^*\leftarrow\{\mu^{\text{test}}=0,\Sigma^{\text{test}}=I\}$}
     \FOR {Time $t = 0$ to $T$}
     \STATE{Solve optimal action $\ba^*_t$ given $p_{\theta^*}$, $q_{\phi^*}$, and MPC \hfill $\triangleright$ see \eqref{eq:mpc}}
     \STATE{Execute action $\ba^*_{t}$}
     \STATE{Record outcome: $\datatest \leftarrow \datatest \cup \{\bs_{t}, \ba^*_{t}, \bs_{t+1}\}$}
     \STATE{Infer variational parameters $\phi^*$ given $\datatest$ \hfill $\triangleright$ see \eqref{eq:latent}\;}
     \ENDFOR
    \end{algorithmic}
    \caption{Model-Based Meta-Reinforcement Learning\\\hspace*{57pt}for Quadcopter Payload Transport}
    \label{alg:full}
\end{algorithm}


\section{Experimental Evaluation}
\label{sect:exp}

\setlength\extrarowheight{2pt}  
\newcolumntype{Y}{>{\centering\arraybackslash}X}
\begin{table*}[t]
\centering
\caption{Comparative evaluation of our method for the tasks of following a circle, square or figure-8 trajectory with either an 18cm or 30cm payload cable length. The table entries specify the average pixel tracking error over 5 trials, with $\infty$ denoting when all trials failed the task by deviating outside of the camera field of view. Note that the cable length was not given to any method \textit{a priori}, and therefore online adaptation was required in order to successfully track the specified path. Our method was able to most closely track all specified paths for all payloads.}
\label{tab:results}
\vspace{-1mm}
\begin{tabularx}{\textwidth}{|l| *{6}{Y|} }
\cline{1-7} \multirow{3}{*}{\,Algorithm} & \multicolumn{6}{c|}{Avg. Tracking Error (pixels) for each Task Path and Payload String Length (cm)} \\
\cline{2-7} & \multicolumn{2}{c|}{Circle} & \multicolumn{2}{c|}{Square} & \multicolumn{2}{c|}{Figure-8} \\
\cline{2-7} & 18 & 30 & 18 & 30 & 18 & 30 \\
\cline{1-7}
\cline{1-7} 
\,Ours (unknown variable)\, & \textbf{23.62}$\pm$\textbf{2.67} & \textbf{24.41}$\pm$\textbf{3.90} & \textbf{23.88}$\pm$\textbf{2.81} & \textbf{26.57}$\pm$\textbf{3.84}  & \textbf{24.67}$\pm$\textbf{1.33} & 29.08$\pm$6.00 \\
\,Ours (known variable) & 31.81$\pm$6.49 & 30.49$\pm$2.65 & 26.37$\pm$3.63 & 31.68$\pm$4.68 & 29.84$\pm$2.84 & \textbf{28.28}$\pm$\textbf{3.76}  \\
\cline{1-7} 
\,MBRL & $\infty$ & $\infty$ & $\infty$ & $\infty$ & $\infty$ & $\infty$  \\
\,MBRL with history & 39.96$\pm$4.40 & 42.36$\pm$2.84 & 32.37$\pm$2.40 & 39.26$\pm$5.16 & 34.17$\pm$1.90 & 41.01$\pm$7.26 \\
\,PID controller & 70.58$\pm$4.01 & 67.98$\pm$2.50 & 65.79$\pm$9.99 & 69.53$\pm$6.85  & 90.15$\pm$10.40 & 86.37$\pm$9.27 \\
\cline{1-7}
\end{tabularx}
\vspace{2mm}
%
\setlength\extrarowheight{2pt}  
\newcolumntype{Y}{>{\centering\arraybackslash}X}
\centering
\caption{Generalization results. Training data consists of 18cm \& 30cm cable lengths augmented with only a few minutes of 24cm data. At test-time, we use 21cm and 27cm cables, to test the model's ability to adapt to new dynamics at intermediate cable lengths.
}
\label{tab:generalize}
\vspace{-1mm}
{\scriptsize
\begin{tabularx}{\textwidth}{|l| *{6}{Y|} }
\cline{1-7} \multirow{3}{*}{\,Algorithm} & \multicolumn{6}{c|}{Avg. Tracking Error (pixels) for each Task Path and Payload String Length (cm)} \\
\cline{2-7} & \multicolumn{2}{c|}{Circle} & \multicolumn{2}{c|}{Square} & \multicolumn{2}{c|}{Figure-8} \\
\cline{2-7} & 21cm & 27cm & 21cm & 27cm & 21cm & 27cm \\
\cline{1-7}
\cline{1-7} 
\,Ours (3 unknown variables)\, & \textbf{16.8}$\pm$\textbf{1.3} & \textbf{21.7}$\pm$\textbf{2.1} & \textbf{24.1}$\pm$\textbf{2.4} & \textbf{24.3}$\pm$\textbf{1.4}  & \textbf{36.2}$\pm$\textbf{2.5} & \textbf{40.1}$\pm$\textbf{3.8} \\
\cline{1-7} 
\,MBRL with history & 21.1$\pm$4.7 & 25.9$\pm$1.6 & 28.3$\pm$2.5 & 37.0$\pm$2.6 & 43.8$\pm$3.7 & 41.9$\pm$3.9 \\
\cline{1-7}
\end{tabularx}
}
\end{table*}

\newcommand{\Qadaptisbetter}{\textbf{Q1}}
\newcommand{\Qmetabetter}{\textbf{Q2}}
\newcommand{\Qknownvsunknown}{\textbf{Q3}}
\newcommand{\Qgeneralization}{\textbf{Q4}}
\newcommand{\Qdifferentiate}{\textbf{Q5}}
\newcommand{\Qmagnetic}{\textbf{Q6}}

We now present an experimental evaluation of our meta-learning approach in the context of quadcopter suspended payload control tasks. Videos and supplementary material are available on our website\footnote{\url{https://sites.google.com/view/ral-meta-rl-for-flight}}.

In these experiments, we aim to answer the following questions:
\begin{enumerate}
    \item[\Qadaptisbetter] Does online adaptation via meta-learning lead to better performance compared to non-adaptive methods?
    \item[\Qmetabetter] How does our meta-learning approach compare to MBRL conditioned on a history of states and actions?
    \item[\Qknownvsunknown] How does our approach with known versus unknown dynamics variables compare?
    \item[\Qgeneralization] Can we \textit{generalize} to payloads that were not seen at training time?
    \item[\Qdifferentiate] Is the test-time inference procedure able to differentiate between different \textit{a priori} unknown payloads?
    \item[\Qmagnetic] Can our approach enable a quadcopter to fulfill a complete payload pick-up, transport, and drop-off task, as well as other realistic payload transportation scenarios?
\end{enumerate}

We evaluated our meta-learning approach with both known variables (\sect{method-known-vars}) and latent variables (\sect{method-latent-vars}), and compared to multiple other approaches, including:
\begin{itemize}
    \item \textit{MBRL}, in which the state consists of only the current payload pixel location and size.
    \item \textit{MBRL with history}, a simple meta-learning approach in which the state consists of the past 8 states and actions concatenated together.
    \item \textit{PID controller}, which consists of three PID controllers, one for each Cartesian velocity command axis. We manually tuned the PID gains by evaluating the performance of the controller on a trajectory following path not used in our experiments for a single payload.
\end{itemize}

\setlength{\tabcolsep}{0em}
\begin{figure}[t]
\centering
\begin{tabular}{cccc}
& Ours (Unknown) & Ours (Known) & MBRL w/ history \\
\rotatebox{90}{\hspace{15pt}Circle} & \includegraphics[angle=0,width=0.15\textwidth,trim={1.5cm 2cm 1.5cm 2cm},clip]{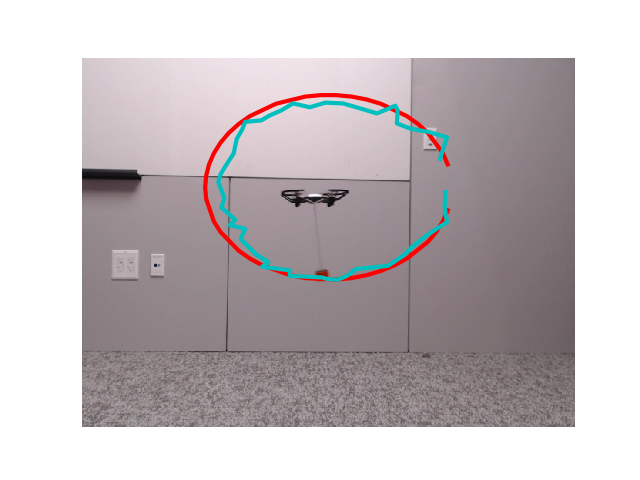} & \includegraphics[angle=0,width=0.15\textwidth,trim={1.5cm 2cm 1.5cm 2cm},clip]{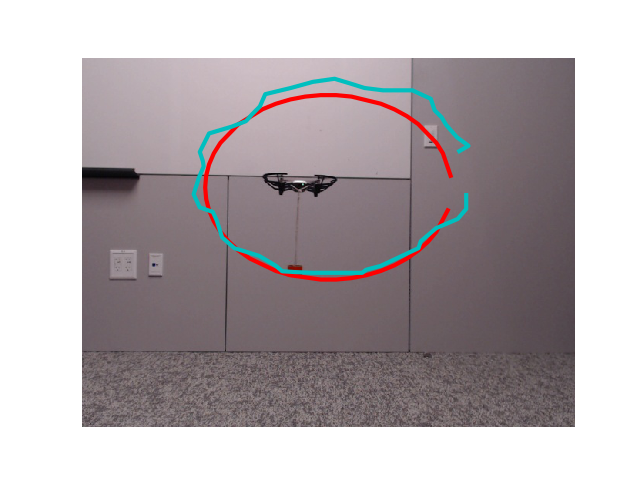} & \includegraphics[angle=0,width=0.15\textwidth,trim={1.5cm 2cm 1.5cm 2cm},clip]{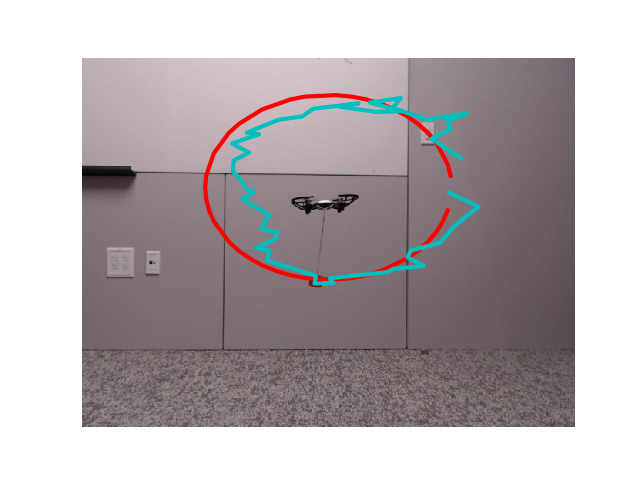} \\
\rotatebox{90}{\hspace{10pt}Square} & \includegraphics[angle=0,width=0.15\textwidth,trim={1.5cm 2cm 1.5cm 2cm},clip]{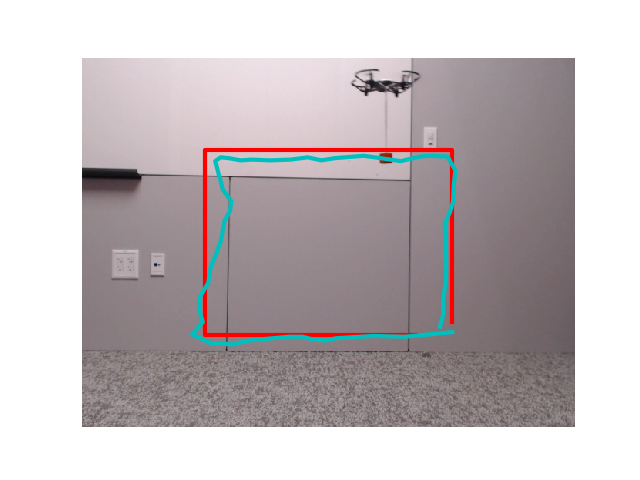} & 
\includegraphics[angle=0,width=0.15\textwidth,trim={1.5cm 2cm 1.5cm 2cm},clip]{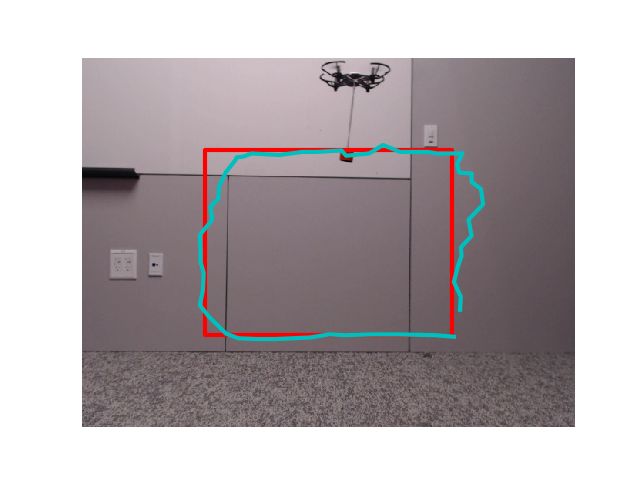} & 
\includegraphics[angle=0,width=0.15\textwidth,trim={1.5cm 2cm 1.5cm 2cm},clip]{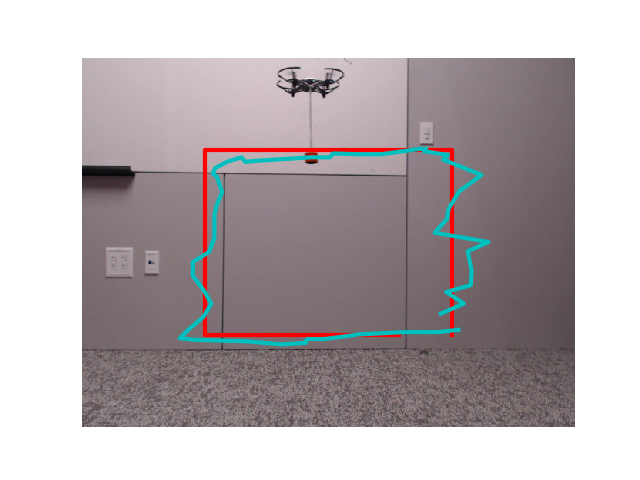} \\
\rotatebox{90}{\hspace{10pt}Figure-8} & \includegraphics[angle=0,width=0.15\textwidth,trim={1cm 1cm 1cm 1cm},clip]{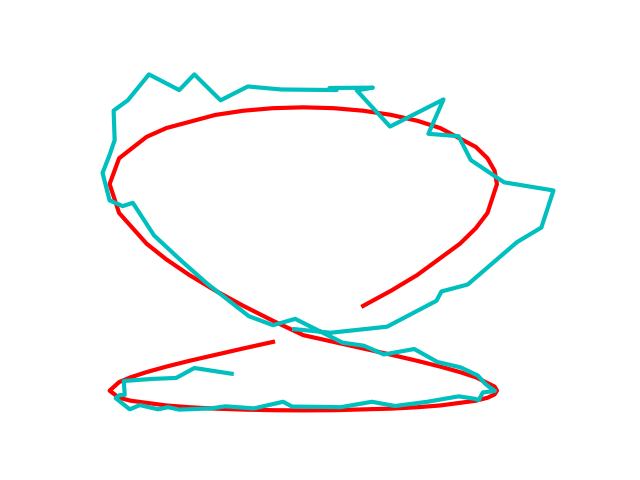} & \includegraphics[angle=0,width=0.15\textwidth,trim={1cm 1cm 1cm 1cm},clip]{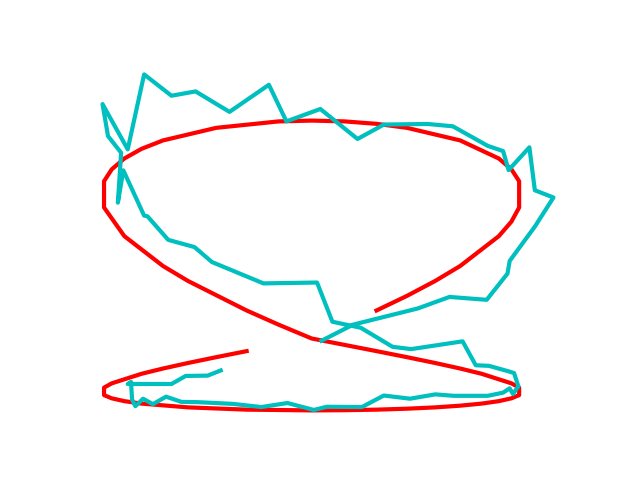} & \includegraphics[angle=0,width=0.15\textwidth,trim={1cm 1cm 1cm 1cm},clip]{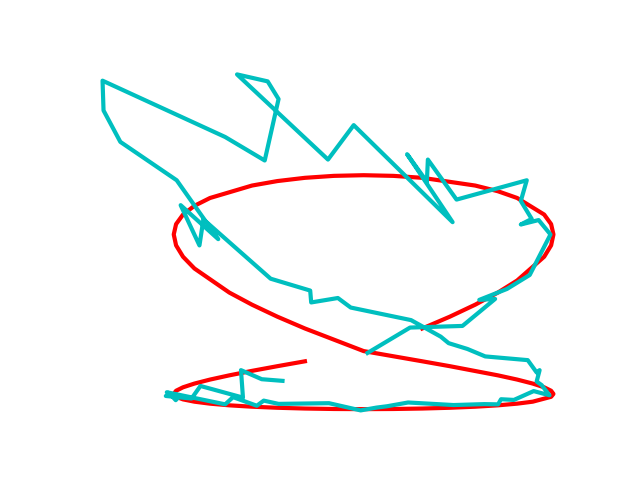} \\
\end{tabular}
\caption{Comparison of our meta-learning approach with unknown and known factors of variation versus model-based reinforcement learning (MBRL) conditioned on a history of states and actions. The tasks are to either follow a circle or square in the image plane, or a figure-8 parallel to the ground. The specified goal paths are colored in red and the path taken by each approach is shown in cyan. Our approaches are better able to adapt online and follow the specified trajectories.}
\label{fig:results-paths}
\vspace{-10pt}
\end{figure}

\subsection{Trajectory Following}
We first evaluate the ability of our method to track specified payload trajectories in isolation, separately from the full payload transportation task. Each task consists of following either a circle or square path (\fig{results-paths}) in the image plane or a figure-8 path parallel to the ground, and with a suspended cable either 18cm or 30cm long. 
Given this single factor of variation, we used a latent variable of dimension one. Although the training data included payloads with these cable lengths, the length was unknown to all methods during test-time experiments.

\tab{results} shows the results for each approach in terms of average pixel tracking error, with visualizations of a subset of the executions shown in \fig{results-paths}. Both the online adaptation methods---our approach and MBRL with history---better track the specified goal trajectories compared to the non-adaptation methods---MBRL and PID controller---which shows that online adaptation leads to better performance (\Qadaptisbetter). Our meta-learning approach also outperforms the other meta-learning method MBRL with history (\Qmetabetter).
Interestingly, our approach with unknown latent variables at training time outperformed our approach with known latent variables (\Qknownvsunknown). 
A possible explanation is that inferring unknown latent variables at training-time captures unspecified types of variation from potentially hard to observe factors, in comparison to specifying observable types of variation (e.g. tether length). In addition, sampling latent variables from a distribution with full support during training could prevent test set latent samples from being out of distribution.
Nevertheless, this shows our approach does not require \textit{a priori} knowledge of latent factors during training to successfully adapt at test time.
We also demonstrate our method's ability to generalize to new payloads not seen during training (\Qgeneralization). 
While we found training using only two string lengths was sufficient to learn to adapt to either dynamics, 
the ability to generalize improves with several example tasks. In our case, learning how string lengths affect the dynamics benefited from a few minutes of data from a third (24cm) string length, allowing us to rapidly interpolate to unseen string lengths of 21cm or 27cm at test-time, shown in \tab{generalize}. 

\begin{figure}[t]
\centering
\begin{tabular}{ccc}
& Known latent & Unknown latent  \\
\rotatebox{90}{\hspace{22pt}String $l = 18$cm}  & \includegraphics[angle=0,width=0.45\columnwidth,height=0.45\columnwidth]{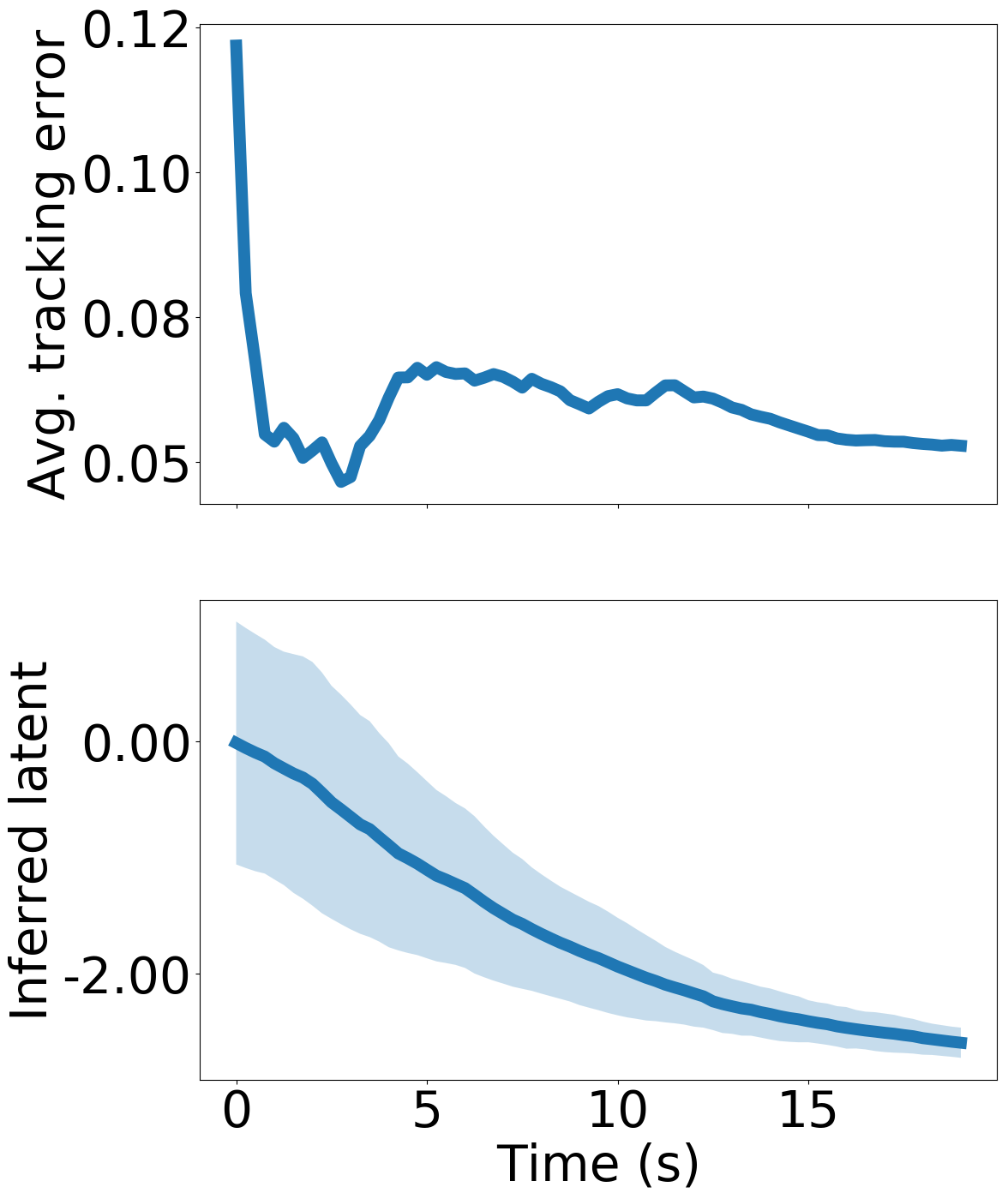} & \includegraphics[angle=0,width=0.45\columnwidth,height=0.45\columnwidth]{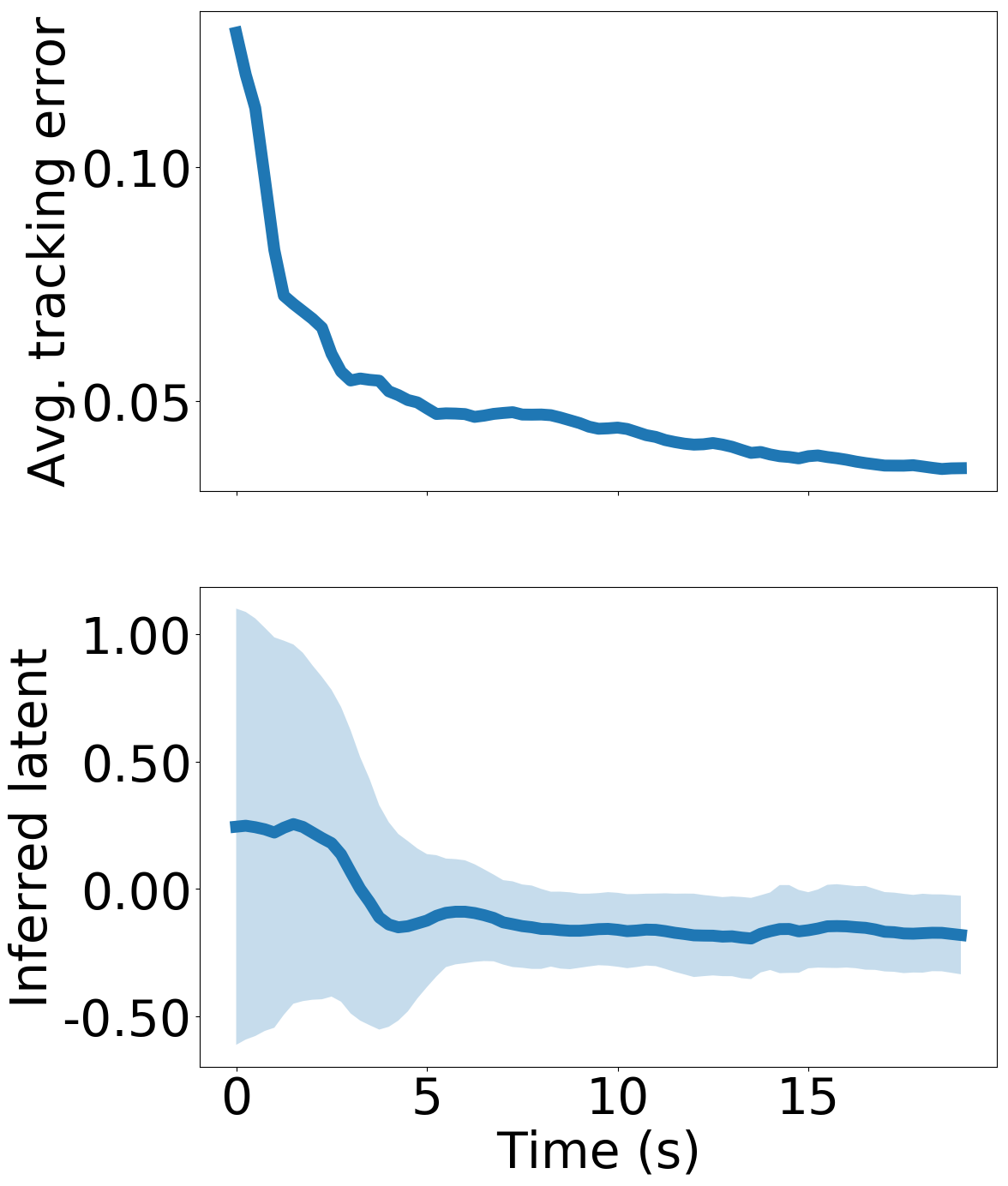} \\
\rotatebox{90}{\hspace{22pt}String $l = 30$cm} & \includegraphics[angle=0,width=0.45\columnwidth,height=0.45\columnwidth]{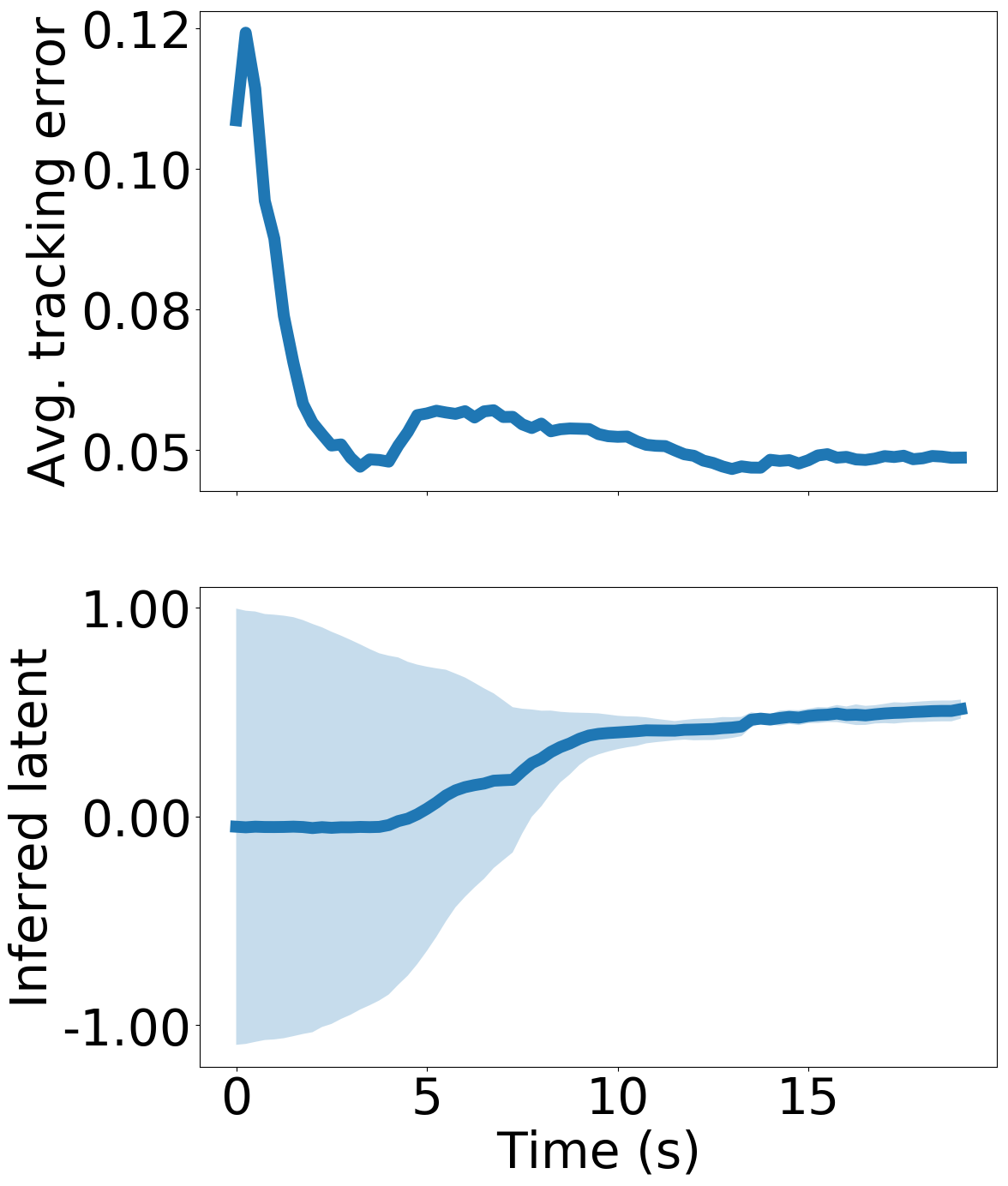} & \includegraphics[angle=0,width=0.45\columnwidth,height=0.45\columnwidth]{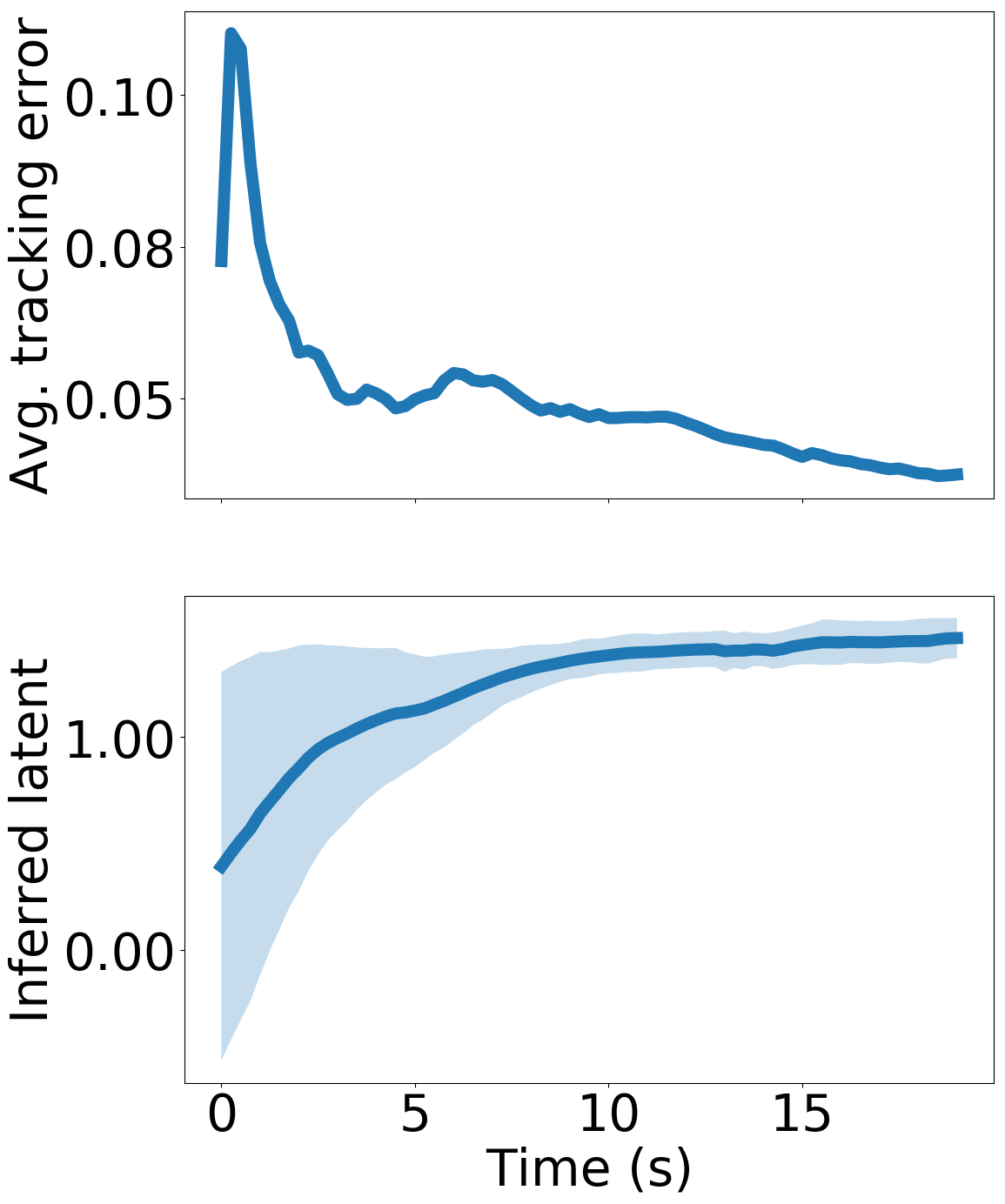} \\
\end{tabular}
\caption{Visualization of the inferred latent variable and tracking error over time for the task of following a figure-8 trajectory. We show our approach trained with known variables (left column) and unknown variables (right column) with either a payload cable length of \SI{18}{\centi\meter} (top row) or \SI{30}{\centi\meter} (bottom row). For all approaches, the inferred latent variable converges as the quadcopter flies and adapts online. The converged final latent values are different depending on the cable length, which shows the online adaptation mechanism is able to automatically differentiate between the different payloads. Furthermore, as the latent value converges, the tracking error also reduces, which demonstrates that there is a correlation between inferring the correct latent variable and the achieved task performance.}
\label{fig:results-ours-test-time}
\end{figure}

\fig{results-ours-test-time} and \fig{results-circle-thumbnail} show the inferred dynamics variable and tracking error while our model-based policy is executing at test time. We observe that the dynamics variable converges to different values depending on the cable length, which shows that our test-time inference procedure is able to differentiate between the dynamics of the two different payloads (\Qdifferentiate). More importantly, as the inferred value converges, our learned model-based controller becomes more accurate and is therefore better able to track the desired path (\Qadaptisbetter).

\subsection{End-to-End Payload Transportation}
\begin{figure*}[t]
\begin{minipage}[c]{\textwidth}
  \begin{minipage}[c]{0.61\textwidth}
    \includegraphics[width=\textwidth]{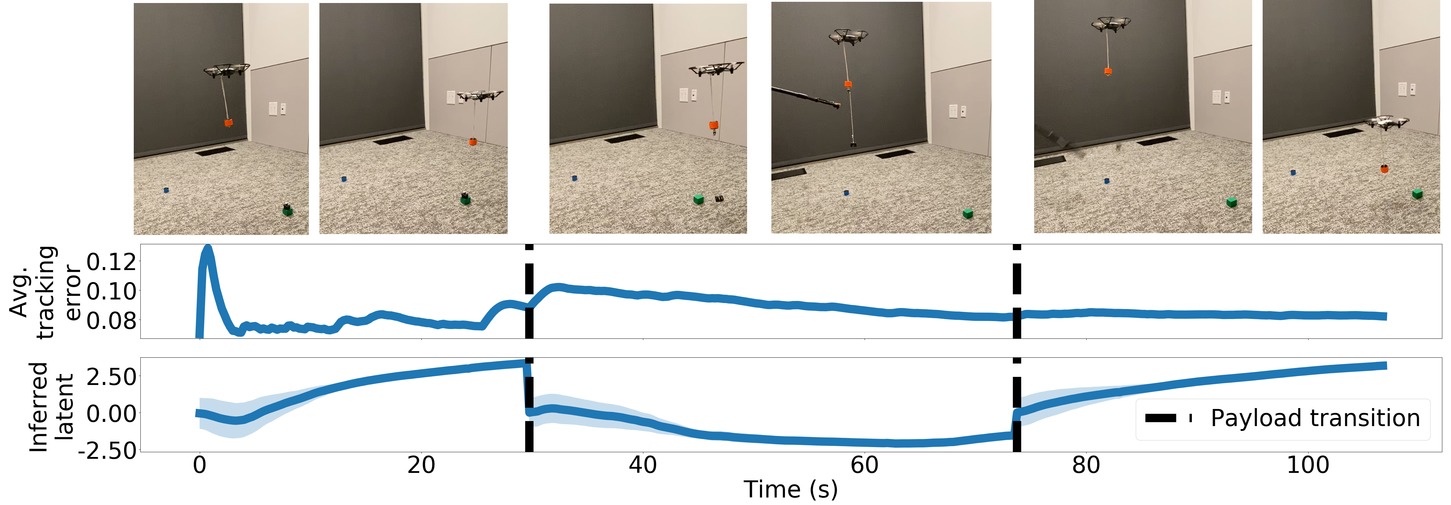}
  \end{minipage}\hfill
  \begin{minipage}[c]{0.37\textwidth}
    \vspace{14mm}
    \caption{Visualization of our approach successfully completing the full quadcopter payload transportation task. The task consists of three phases: the quadcopter before picking up the payload; while the payload is in transit to the goal; and after the payload is dropped off. Our approach continuously adapts the latent dynamics variable online using the current test-time dataset, flushing the test-time dataset each time the quadcopter transitions between phases, shown by vertical black lines. Note the inferred latent variable is the same for when no payload is attached, but different when the payload is attached, which demonstrates that our inference procedure successfully infers the latent variable depending on the payload. Within each phase, the tracking error also reduces over time, which shows that our online adaptation mechanism improves closed-loop performance.}
    \label{fig:results-magnetic-thumbnail}
  \end{minipage}
  \begin{minipage}[c]{0.61\textwidth}
    \vspace{-12mm}
    \includegraphics[width=\textwidth]{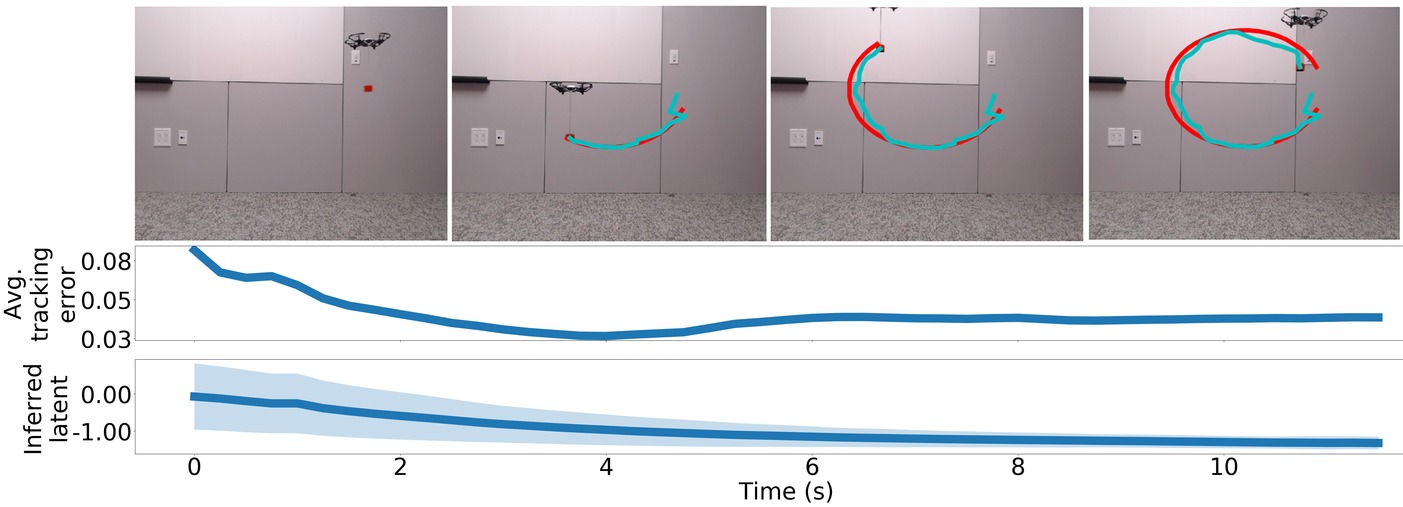}
  \end{minipage}\hfill
  \begin{minipage}[c]{0.37\textwidth}
   \vspace{-1mm}
    \caption{As the quadcopter follows the circle trajectory using our model-based controller, our approach adapts online to the \textit{a priori} unknown payload by inferring the latent value which maximizes the dynamics models accuracy. This online adaptation reduces the tracking error as the quadcopter flies, enabling the quadcopter to successfully complete the task.}
    \label{fig:results-circle-thumbnail}
  \end{minipage}
\end{minipage}

\vspace{-6mm}
\end{figure*}

We also evaluated our approach on an end-to-end payload transportation task (\fig{results-magnetic}), in which the quadcopter must follow a desired trajectory to the payload, attach to the payload using a magnet, lift the payload and transport it along a specified trajectory to the goal location, drop off the payload, and then follow a trajectory back to the start location.
\fig{results-magnetic-thumbnail} shows 
our approach successfully completes the full task (\Qmagnetic) due to our online adaptation mechanism (\Qadaptisbetter), which enables the drone to follow trajectories better and pick up the payload by automatically inferring whether the payload is attached or detached (\Qdifferentiate). Furthermore, the continuous nature of this task highlights the importance of online adaptation: each time the quadcopter transitions between transporting a payload and not transporting a payload, the quadcopter must re-adapt online to be able to successfully follow the specified trajectories.

\subsection{Additional Use Cases}
In addition to enabling trajectory following and end-to-end payload transportation, we demonstrated our approach to transport a suspended payload (\Qmagnetic): towards a moving target, around an obstacle by following a predefined path, and along trajectories dictated using a ``wand''-like interface (see videos and website).


\section{Discussion \& Conclusion} 
\label{sect:discussion}

We presented a meta-learning approach for model-based reinforcement learning that enables a quadcopter to adapt to various payloads in an online fashion with minimal state information. At the core of our approach is a deep neural network dynamics model that learns to predict how the quadcopter's actions affect the flight path of the payload. We augment this dynamics model with stochastic latent variables, which represent unknown factors of variation in the task. These latent variables are trained to improve the accuracy of the dynamics model and be amenable for fast online adaptation. Our experiments demonstrate that the proposed training and online adaptation mechanisms improve performance for real-world quadcopter suspended payload transportation tasks compared to other adaptation approaches.

{
\bibliographystyle{plainnat}
\small
\bibliography{references}
}

\end{document}

%% file: preamble.tex
\usepackage{algorithm}
\usepackage[noend]{algorithmic}
\usepackage{amsmath}
\usepackage{amssymb}
\usepackage{color}
\usepackage{xcolor}
\usepackage{graphicx}
\usepackage{dblfloatfix}
\usepackage{multicol}
\usepackage{multirow}
\usepackage[numbers]{natbib}
\usepackage{subcaption}
\usepackage{tabularx}
\usepackage{times}
\usepackage{siunitx}
\usepackage{url}

\newcommand{\bs}{\mathbf{s}}
\newcommand{\ba}{\mathbf{a}}

\newcommand{\bz}{\mathbf{z}}
\newcommand{\bztest}{\bz^{\text{test}}}
\newcommand{\bzall}{\bz_{1:K}}
\newcommand{\datatrain}{\mathcal{D}^{\text{train}}}
\newcommand{\datatest}{\mathcal{D}^{\text{test}}}
\newcommand{\task}{\mathcal{T}}
\newcommand{\setK}[1]{#1_{1:K}}
\newcommand{\phitest}{\phi^{\text{test}}}

\newcommand{\algo}[1]{Algorithm~\ref{algo:#1}}
\newcommand{\fig}[1]{Figure~\ref{fig:#1}}
\newcommand{\sect}[1]{\S\ref{sect:#1}}
\newcommand{\tab}[1]{Table~\ref{tab:#1}}

\DeclareMathOperator*{\argmax}{argmax}

\usepackage{tikz}
\usetikzlibrary{arrows,shapes,backgrounds,patterns,fadings,decorations.pathreplacing,decorations.pathmorphing}
\tikzset{>=stealth'}
\usetikzlibrary{arrows}
\usetikzlibrary{bayesnet}
\newcommand{\coloryellow}[0]{yellow!20}
\newcommand{\colordarkgrey}[0]{black!10}

\newcommand{\colorlightgrey}[0]{black!50}

\newcommand{\lightgrey}[1]{\textcolor{\colorlightgrey}{#1}}%
\usepackage[skins]{tcolorbox}

%% file: tikz/flow.tex
\pgfdeclarelayer{background}
\pgfdeclarelayer{foreground}
\pgfsetlayers{background,main,foreground}

\tikzstyle{arrownode}  = [above, text width=5em]
\tikzstyle{emptynode}  = [text centered, minimum height=2.5em]
\tikzstyle{bluenode}   = [emptynode, draw, fill=blue!20, text width=2.5em]
\tikzstyle{photonode}   = [emptynode, draw, fill=black!20, text width=2.5em]
\tikzstyle{photonodehigh}   = [emptynode, draw, fill=black!20, text width=2.5em, text height=5em]
\tikzstyle{rednode}    = [emptynode, draw, circle, fill=red!25, text width=1.5em]
\tikzstyle{bigrednode} = [emptynode, draw, fill=red!12,  text width=6em, text height=11em, rounded corners]
\tikzstyle{medrednode} = [emptynode, draw, fill=red!12,  text width=3.6em, text height=5em, rounded corners]
\tikzstyle{yellownode} = [emptynode, fill=\coloryellow, text width=.5em, text height=.3em, minimum height=.3em]

\def\blockdist{1.94}
\def\shifttop{0.8}
\def\shiftmid{-0.1}
\def\shiftbot{-1}
\def\shifttikz{0mm}

\begin{figure*}[t]
\begin{tikzpicture}
    \hspace{\shifttikz}
    
    \node (infer) [bigrednode] {};


    \def\datashift{3.2}
    \path (infer)+(-\datashift,\shifttop) node (data1) [bluenode]  {$\datatrain_1$};
    \path (infer)+(-\datashift,\shiftmid) node (data_) [emptynode] {$\vdots$};
    \path (infer)+(-\datashift,\shiftbot) node (dataK) [bluenode]  {$\datatrain_K$};
    \path (data1.north)+(0.0,0.5)         node (data)              {Data $\datatrain$};
    
    \def\phishift{0.5}
    \path (infer)+(-\phishift,\shifttop) node (phi1) [rednode]   {$\phi_1$};
    \path (infer)+(-\phishift,\shiftmid) node (phi_) [emptynode] {$\vdots$};
    \path (infer)+(-\phishift,\shiftbot) node (phiK) [rednode]   {$\phi_K$};
    \path (phi1.north)+(\phishift,0.5)   node                    {Meta-Training};

    \path (infer)+(0.5,\shiftmid) node (theta) [rednode] {$\theta$};

    \path (data1)+(-\blockdist,0) node (system1) [photonode, fill overzoom image=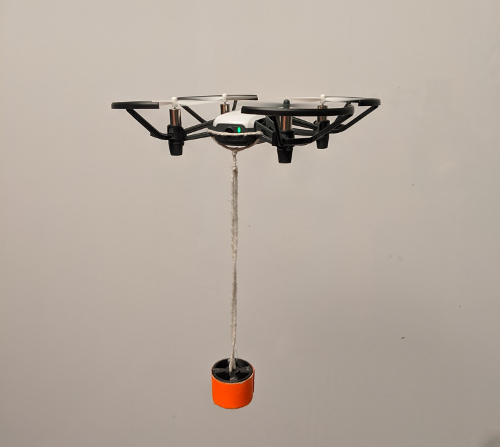]  {};
    \path (data_)+(-\blockdist,0) node (system_) [emptynode] {$\vdots$};
    \path (dataK)+(-\blockdist,0) node (systemK) [photonode, fill overzoom image=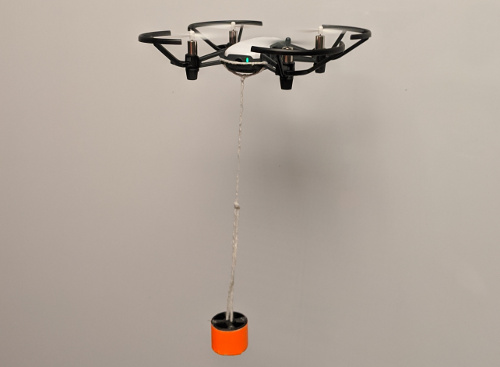]  {};
    
    \path (system1)+(-\blockdist,0) node (policy1) [photonode, fill overzoom image=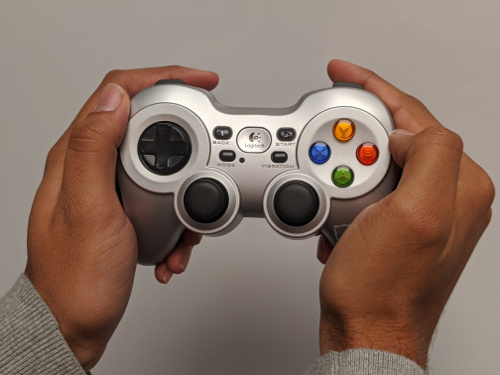]  {};
    \path (system_)+(-\blockdist,0) node (policy_) [emptynode] {$\vdots$};
    \path (systemK)+(-\blockdist,0) node (policyK) [photonode, fill overzoom image=figures/joystick.jpg]  {};
    
    \path (policy1)+(0,0.65) node {\lightgrey{\small{Teleop}}};
    \path (system1)+(0,0.65) node {\lightgrey{\small{System}}};
    \path (policy1)+(-1.05,0) node [rotate=90] {\lightgrey{\small{Task 1}}};
    \path (policy_)+(-1.05,0) node [rotate=90] {\lightgrey{\small{...}}};
    \path (policyK)+(-1.05,0) node [rotate=90] {\lightgrey{\small{Task K}}};
    
    \path [draw, ->] (data1) -- node [above] {\;\;$\bs,\!\ba,\!\bs'$} (infer.west |- data1);
    \path [draw, ->] (dataK) -- node [above] {\;\;$\bs,\!\ba,\!\bs'$} (infer.west |- dataK);
    \path [draw, ->] (system1) -- node [above] {$\bs$\;\;} (data1.west |- system1);
    \path [draw, ->] (systemK) -- node [above] {$\bs$\;\;} (dataK.west |- systemK);
    \path [draw, ->] (policy1) -- node [above] {$\ba$} (system1.west |- policy1);
    \path [draw, ->] (policyK) -- node [above] {$\ba$} (systemK.west |- policyK);
    \path [draw, ->] (policy1.east)+(0.5,0) |- +(0.5,-0.75) -| (data1.south west); 
    \path [draw, ->] (policyK.east)+(0.5,0) |- +(0.5,-0.75) -| (dataK.south west);
    
    \path (policy1.north)+(0.4,1.2) node (training) {\lightgrey{Training Phase}};

    \begin{pgfonlayer}{background}
        \hspace{\shifttikz}
        \path (policy1.north west)+(-0.2,1.0) node (a) {}; 
        \path (infer.south east)+(+0.2,-0.2) node (b) {};   
        \path[fill=yellow!20,rounded corners, draw=\colorlightgrey, dashed] (a) rectangle (b);
        \path (data1.north west)+(-0.3,0.75) node (a) {}; 
        \path (dataK.south east)+(+0.3,-0.45) node (b) {}; 
        \path[fill=blue!10,rounded corners, draw=\colorlightgrey, dashed] (a) rectangle (b);
    \end{pgfonlayer}
    
    
    \path (infer)+(2.6,\shiftmid)  node (systemT) [photonodehigh, fill overzoom image=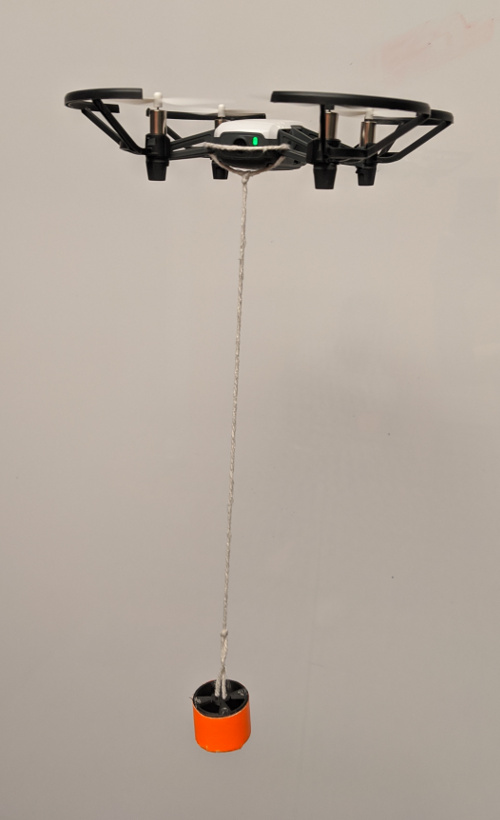] {};         
    \path (systemT)+(\blockdist,0) node (dataT)   [bluenode]  {$\datatest$};    
    \path (dataT)+(2.5,0)   node (inferT)    [medrednode]   {};         
    \path (inferT)+(0,0)   node (phiT)    [rednode]   {$\phitest$};         
    \path (phiT)+(\blockdist,0)    node (policyT)    [bluenode]   {MPC};         
    \path (phiT.north)+(0,0.3)   node                    {Meta-Test};
    
    \path [draw, ->] (systemT) -- node [above] {$\bs$} (dataT.west |- systemT);
    \path [draw, ->] (dataT) -- node [above] {$\bs,\!\ba^{\!*}\!,\!\bs'$} (inferT.west |- dataT);
    \path [draw, ->] (inferT) -- node [above] {$\phi^*$} (policyT.west |- inferT);
    \def\actionshift{1.35}
    \draw[->] (policyT.south) |- +(0,-\actionshift) |- +(-3.,-\actionshift) node [above] {$\ba^*$} -| (systemT.south); 
    \draw[->] (dataT.south)+(0,-\actionshift) |- +(0.,0); 
    \path (dataT)+(0,-1.2) node [yellownode] {}; 
    \path [draw, ->] (systemT.east)+(0.5,0) |- +(0.5,-1.2) -| (policyT.south west);
    \path (systemT.north)+(0.1,0.6) node (test) {\lightgrey{Test Phase}};

    \begin{pgfonlayer}{background}
        \hspace{\shifttikz}
        \path (systemT.north west)+(-0.2,0.4) node (a) {}; 
        \path (policyT.south east)+(+0.2,-1.7) node (b) {};   
        \path[fill=\coloryellow,rounded corners, draw=\colorlightgrey, dashed] (a) rectangle (b);
    \end{pgfonlayer}

    \draw [->] (infer.58) -| (policyT.north);
    \draw [->] (infer.58) -- node [arrownode] {\text{dynamics model parameters}\;\,$\theta^*$} + (5.5,0)  -| (inferT.north);

\end{tikzpicture}
\caption{System diagram of our meta-learning for model-based reinforcement learning algorithm. In the training phase, we first gather data by manually piloting the quadcopter along random trajectories with $K$ different payloads, and saving the data into a single dataset $\datatrain$ consisting of $K$ separate training task-specific datasets $\datatrain\doteq\setK{\datatrain}$. We then run meta-training to learn the shared dynamics model parameters $\theta$ and the adaptation parameters $\phi_{1:K}$ for each payload task. At test time, using the learned dynamics model parameters~$\theta^*$, the robot infers the optimal latent variable~$\phi^*$ online using all of the data~$\datatest$ from the current task. The dynamics model, parameterized by $\theta^*$ and $\phi^*$, is used by a model-predictive controller (MPC) to plan and execute actions that follow the specified path. As the robot flies, it continues to store data, infer the optimal latent variable parameters, and perform planning in a continuous loop until the task is complete.}
\label{fig:flow}

\vspace{-5pt}
\end{figure*}

%% file: tikz/graphs.tex
\begin{figure}[t]
\centering
\begin{subfigure}[t]{\linewidth}
    \centering
    \resizebox{0.8\linewidth}{!}{
    \begin{tikzpicture}[->,>=stealth',scale=1, transform shape]
    \node [matrix,matrix anchor=mid, column sep=12pt, row sep=12pt, ampersand
    replacement=\&,nodes={circle,thick,minimum size=1.0cm}] {
    \node[draw, fill=\colordarkgrey] (a0)   {$\ba_{t}$};   \&
    \node[draw, fill=\colordarkgrey] (s0)   {$\bs_{t}$};   \&
    \node[draw, fill=\colordarkgrey] (s1)   {$\bs_{t+1}$}; \&
    \node[draw, fill=\colordarkgrey] (zk)   {$\bz_{k}$};   \&
    \node[]                          (pk)   { };           \&
    \node[draw, fill=\colordarkgrey] (th)   {$\theta$};    \\
    };
    \path (pk)+(0.1,0) node (cK) {};  
    \draw [->,out=145,in=38] (th) to (s1) ;
    \draw [->,out=38,in=142] (a0) to (s1) ;
    \draw [->] (s0) to (s1) ;
    \draw [->] (zk) to (s1) ;
    \plate [inner sep=3mm, yshift=0.4mm] {plateT} {(a0)(s1)} {}; 
    \plate [inner sep=5mm, yshift=0.4mm] {plateK} {(a0)(cK)} {}; 
    \path (s1)+(0,-0.75) node {\small{$t\in[T]$}};  
    \path (pk)+(0,-0.88) node {\small{$k\in[K]$}};  
    \end{tikzpicture}
    }
    \caption{\textbf{Training}-time payloads with \textbf{known} factors of variation $\bz_k$.}
    \label{fig:graphKnownZ}
\end{subfigure}
\vspace{2mm}

\begin{subfigure}[t]{\linewidth}
    \centering
    \resizebox{0.8\linewidth}{!}{
    \begin{tikzpicture}[->,>=stealth',scale=1, transform shape]
    \node [matrix,matrix anchor=mid, column sep=12pt, row sep=12pt, ampersand
    replacement=\&,nodes={circle,thick,minimum size=1.0cm}] {
   \node[draw, fill=\colordarkgrey]  (a0)    {$\ba_{t}$};  \&
    \node[draw, fill=\colordarkgrey] (s0)   {$\bs_{t}$};   \&
    \node[draw, fill=\colordarkgrey] (s1)   {$\bs_{t+1}$}; \&
    \node[draw]                      (zk)   {$\bz_{k}$};   \&
    \node[draw, fill=\colordarkgrey] (pk)   {$\phi_{k}$};  \&
    \node[draw, fill=\colordarkgrey] (th)   {$\theta$};   \\
    };
    \path (pk)+(0.1,0) node (cK) {};  
    \draw [->,out=145,in=38] (th) to (s1) ;
    \draw [->,out=38,in=142] (a0) to (s1) ;
    \draw [->] (s0) to (s1) ;
    \draw [->] (zk) to (s1) ;
    \draw [->] (pk) to (zk) ;
    \plate [inner sep=3mm, yshift=0.4mm] {plateT} {(a0)(s1)} {}; 
    \plate [inner sep=5mm, yshift=0.4mm] {plateK} {(a0)(cK)} {}; 
    \path (s1)+(0,-0.75) node {\small{$t\in[T]$}};  
    \path (pk)+(0,-0.88) node {\small{$k\in[K]$}};  
    \end{tikzpicture}
    }
    \caption{\textbf{Training}-time payloads with \textbf{unknown} factors of variation $\bz_k$.}
    \label{fig:graphUnknownZ}
\end{subfigure}
\vspace{2mm}

\begin{subfigure}[t]{\linewidth}
    \centering
    \resizebox{0.8\linewidth}{!}{
    \begin{tikzpicture}[->,>=stealth',scale=1, transform shape]
    \node [matrix,matrix anchor=mid, column sep=12pt, row sep=12pt, ampersand
    replacement=\&,nodes={circle,thick,minimum size=1.0cm}] {
   \node[draw, fill=\colordarkgrey]  (a0)    {$\ba^*_{t}$};  \&
    \node[draw, fill=\colordarkgrey] (s0)   {$\bs_{t}$};   \&
    \node[draw, fill=\colordarkgrey] (s1)   {$\bs_{t+1}$}; \&
    \node[draw]                      (zk)   {$\bztest$};   \&
    \node[draw, fill=\colordarkgrey] (pk)   {$\phitest$};  \&
    \node[draw, fill=\colordarkgrey] (th)   {$\theta^*$};   \\
    };
    \draw [->,out=145,in=38] (th) to (s1) ;
    \draw [->,out=38,in=142] (a0) to (s1) ;
    \draw [->] (s0) to (s1) ;
    \draw [->] (zk) to (s1) ;
    \draw [->] (pk) to (zk) ;
    \plate [inner sep=3mm, yshift=0.4mm] {plateT} {(a0)(s1)} {}; 
    \path (s1)+(0,-0.75) node {\small{$t\in[T]$}};  
    \end{tikzpicture}
    }
    \caption{\textbf{Test}-time payload with unknown factors of variation $\bztest$.}
    \label{fig:graphTest}
    \vspace{1mm}
\end{subfigure}
\caption{Probabilistic graphical models of the drone-payload system dynamics. At each time step $t$, the system state evolves as a function of the current state $\bs_{t}$, action $\ba_{t}$, function parameters $\theta$, and dynamics variable $\bz_k$ which encodes the $k$'th payload's idiosyncrasies. Shaded nodes are observed. At training time, the factors of variation between payloads might be known (Fig.~\ref{fig:graphKnownZ}) or unknown (Fig.~\ref{fig:graphUnknownZ}) 
while training model's parameters $\theta$. Regardless of the training regime, test-time $\bztest$ is always unknown (Fig.~\ref{fig:graphTest}), which we infer given the trained (fixed) model parameters $\theta^*$.}
\label{fig:graphs}
\end{figure}